\ifdefined\pdfminorversion\pdfminorversion=7\fi
\PassOptionsToPackage{authoryear,sort&compress,round}{natbib}
\documentclass[11pt,a4paper,onecolumn]{thuc3i}
\usepackage{l3draw}
\usepackage{natbib}
\usepackage{wrapfig}
\usepackage{flafter}
\graphicspath{{figs/}}

\ExplSyntaxOn
\box_new:N \l_research_question_box
\NewDocumentCommand{\researchquestionbox}{m}
  {
    \group_begin:
    \hbox_set:Nn \l_research_question_box
      { \parbox{\dimexpr\linewidth-16pt\relax}{\raggedright\itshape #1} }
    \draw_begin:
      \draw_baseline:n { 0pt }
      \color_fill:nn { rgb } { 0.9960784314, 0.9960784314, 0.9529411765 }
      \color_stroke:n { black!60 }
      \draw_linewidth:n { 0.8pt }
      \draw_path_corner_arc:nn { 8pt } { 8pt }
      \draw_path_rectangle:nn
        { 0.4pt, -\box_dp:N \l_research_question_box - 7.6pt }
        { \linewidth - 0.8pt,
          \box_ht:N \l_research_question_box + \box_dp:N \l_research_question_box + 15.2pt }
      \draw_path_use_clear:n { fill, stroke }
      \color_select:n { black }
      \draw_transform_shift:n { 8pt, 0pt }
      \draw_box_use:N \l_research_question_box
    \draw_end:
    \group_end:
  }
\ExplSyntaxOff

\setheadertext{LatentLoop}
\title{LatentLoop: Scaling Looped MoE Models with Asymmetric Compute Allocation}

\author[1,2,*]{Zefeng~He}
\author[1,2,*]{Yunkai~Chai}
\author[2,3,*]{Xiaoyang~Chen}
\author[2,*]{Tong~Zhu}
\author[2]{Zongliang~Wu}
\author[2]{Yun~Luo}
\author[2,4,$\dagger$]{Yafu~Li}
\author[5]{Yu~Cheng}
\affil[1]{SJTU}
\affil[2]{Shanghai AI Lab}
\affil[3]{USTC}
\affil[4]{CUHK}
\affil[5]{NTU}
\role[*]{Core contributors}
\role[\dagger]{Corresponding author}
\renewcommand{\Resourcefont}{\normalfont\fontsize{10}{12}\selectfont}
\resource{}{%
  \href{https://latentloop-project.github.io/}{\raisebox{-0.12em}{\includegraphics[height=1.05em]{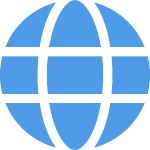}}~Project Page}%
  \hspace{1.5em}%
  \href{https://github.com/lcqysl/LatentLoop}{\raisebox{-0.12em}{\includegraphics[height=1.05em]{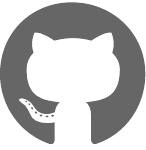}}~Code}%
}

\begin{abstract}
Looped Mixture-of-Experts (MoE) models combine parameter reuse with sparse computation to improve performance under matched budgets.
Prior work demonstrates that different loop visits can play different roles even when they share parameters.
To probe the contributions of different loop visits, we learn separate residual coefficients in a model that applies a shared block twice, observing a smaller scale in the first visit.
Fixed coefficient experiments show that reducing the first visit's scale preserves or improves performance, whereas the same reductions to the second hurt it.
This asymmetry motivates us to explore asymmetric compute allocation.
We propose \textbf{LatentLoop}, which compresses the first visit's sequence, retains full resolution in the second, and reallocates saved computation to increase the hidden size.
We further learn how strongly the first visit updates the full sequence, and share MoE weights across visits while using separate attention parameters to accommodate their different sequence resolutions.
Across 15 configurations spanning 1.3B--7B parameters, with the largest budget exceeding $10^{21}$ FLOPs, LatentLoop consistently lowers validation NLL. At a fitted NLL of 1.988, predicted compute savings are 19.0\% and 7.9\% relative to vanilla MoE and the looped baseline. It also improves average scores on nine benchmarks at all sizes in 0-shot and 5-shot settings. These results suggest asymmetric compute allocation is a promising direction for looped MoE models.
\end{abstract}

\begin{document}
\maketitle
\section{Introduction}
\label{sec:introduction}
Looped Transformers~\citep{dehghani2018universal,bai2019deep,zhu2025scaling,prairie2026parcae,geiping2026scaling} increase effective depth by repeatedly applying shared layers, offering a way to improve language modeling and reasoning without storing a separate set of weights for each visit. However, each visit still requires computation, so gains at a fixed parameter count do not establish an advantage under equal resource budgets. Recent work~\citep{lee2026sparse,chen2026loopmoe,chen2026scaling,wang2026smelt} combines looping with Mixture-of-Experts (MoE)~\citep{fedus2022switch} architectures, allowing stored parameter capacity to be adjusted separately from active computation. In particular, SMELT~\citep{wang2026smelt} shows that looping can improve performance under matched resource budgets. We study how to further improve looped MoE performance while matching parameter count, KV cache, FLOPs per step, and total training FLOPs within 1\%.

We start from SMELT's looped MoE recipe, which applies the middle half of the layers twice using shared parameters. We first examine how the two visits update the model's representations. Prior work suggests that different visits can play different roles even when they share parameters~\citep{geiping2026scaling,zhu2025scaling,wang2026smelt,blayney2026mechanistic}. To probe this difference, we introduce a learnable residual coefficient for each visit and optimize both jointly with the model weights. At both model sizes we study, the first visit learns a substantially smaller coefficient, while the second remains close to one. Experiments with fixed coefficients reinforce this asymmetry: reducing the first visit's residual scale preserves or improves validation NLL, whereas the same reductions to the second visit increase it. This asymmetry in the two visits' preferred residual update scales motivates a further question: \emph{should we allocate different amounts of computation to the two visits?}

Motivated by these findings, we explore asymmetric compute allocation: reducing computation in the first visit and reallocating the saved budget elsewhere. Among several initial approaches, sequence compression gives the lowest validation NLL. We therefore introduce \textbf{LatentLoop}, whose first visit processes a compressed latent sequence while its second operates at full resolution, and reallocate the saved computation to model width. This asymmetry raises two design questions: how to calibrate latent updates written back to full resolution states, and whether the different resolutions affect parameter sharing. Guided by the residual scaling observations, we pair mean aggregation with learnable writeback coefficients to calibrate latent increments with causal alignment. Different sequence resolutions may favor different attention and MoE computations. However, untying MoE blocks would forgo most parameter reuse because they contain most of the model’s parameters. We therefore untie only attention blocks and offset the added parameters by slightly reducing the stored expert count, keeping the active count unchanged.

We evaluate LatentLoop across 15 matched budget configurations spanning 1.3B--7B parameters, with training compute exceeding $10^{21}$ FLOPs. LatentLoop consistently achieves lower validation NLL than both MoE Base and MoE Loop at all 15 configurations. At a fitted NLL of approximately 1.988, the scaling curves predict that LatentLoop requires 19.0\% less training computation than MoE Base and 7.9\% less than MoE Loop, with all compared allocations lying within the observed ranges. We further evaluate nine downstream benchmarks at the largest token budget for each size, using both 0-shot and 5-shot settings. LatentLoop achieves the highest average score in all six size and shot settings, showing that its validation loss gains also carry over to downstream performance. Ablations at the 3B/40B reference budget show the benefits of writeback scale calibration and selective parameter sharing, while analyses of the learned writeback coefficients reveal update scales well below one, echoing the earlier residual scaling observations.

Our main contributions are:
\begin{itemize}
    \item We identify an asymmetry in the two visits' preferred residual scales through learnable coefficients and fixed coefficient experiments. Reducing the first visit's scale can preserve or improve performance, whereas the same reductions to the second hurt performance.
    \item We propose LatentLoop, which compresses the first visit's sequence and reallocates saved computation to model width under matched resource budgets. We introduce learnable writeback coefficients to calibrate latent updates and selectively share MoE routers and experts while keeping attention blocks independent.
    \item We demonstrate lower validation NLL than MoE Base and MoE Loop across all 15 budget configurations, along with higher average downstream scores at all three model sizes in both 0-shot and 5-shot settings. Our experiments suggest that asymmetric compute allocation is a promising direction for improving looped MoE models.
\end{itemize}

\begin{figure}[!t]
    \centering
    \includegraphics[width=\textwidth]{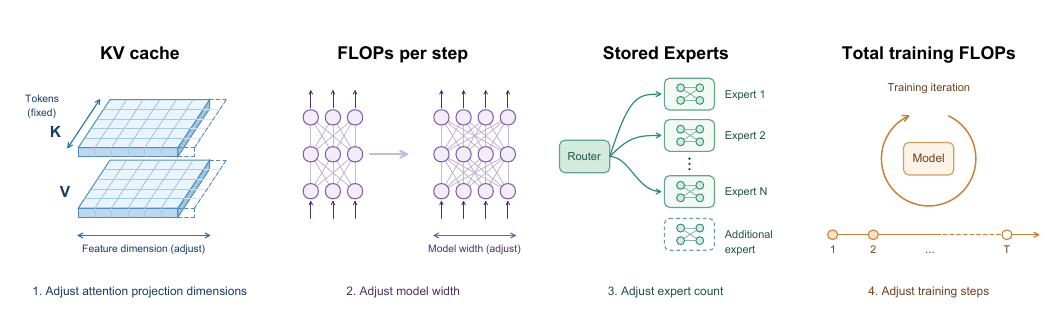}
    \caption{The four steps of budget matching. We adjust attention projection dimensions, model width, the number of stored experts, and training steps in this order.}
    \label{fig:budget_matching}
\end{figure}

\section{Preliminaries}
\label{sec:preliminaries}

We follow the looped MoE architecture and budget matching framework of SMELT~\citep{wang2026smelt}. Our comparisons control stored parameters, KV cache, FLOPs per step, and total training FLOPs, as detailed below.

\subsection{Looped MoE Architecture}
\label{sec:loop_architecture}

We adopt SMELT's basic looped architecture as our starting point. A Transformer with $L$ stored layers is divided into an entry block, a middle block, and an exit block, containing $L/4$, $L/2$, and $L/4$ layers, respectively. The entry and exit blocks are each executed once, while the middle block is visited twice, giving $3L/2$ layer executions. Both visits share all parameters in the middle block, including attention blocks, routers, and experts. We set both residual coefficients to one as a reference for studying residual scaling across visits.

\subsection{Budget Matching}
\label{sec:budget_matching}

\begin{wraptable}{R}{0.48\textwidth}
    \centering
    \small
    \setlength{\tabcolsep}{4pt}
    \renewcommand{\arraystretch}{1.12}
    \begin{tabularx}{\linewidth}{@{}Xrr@{}}
        \toprule
        \multicolumn{1}{@{}c}{\bfseries\boldmath Budget} & \multicolumn{1}{c}{\bfseries\boldmath SMELT} & \multicolumn{1}{c@{}}{\bfseries\boldmath This work} \\
        \midrule
        Backbone parameters & 1.0 \% & 0.6 \%\\
        KV cache & 3.6 \%& 0.8 \%\\
        FLOPs per token & 3.9 \%& 0.9 \%\\
        \bottomrule
    \end{tabularx}
    \caption{Maximum absolute budget deviations.}
    \label{tab:budget_matching_accuracy}
    \end{wraptable}

For a matched comparison, we first adjust the attention projection dimensions to match KV cache, then the model width to match computation per step, and next the number of stored experts to match the parameter budget. Finally, we choose an integer number of optimizer steps to match total training FLOPs. Figure~\ref{fig:budget_matching} summarizes these four steps. The number of query heads, KV groups, and active experts is held fixed within each matched comparison. Parameter matching uses the backbone excluding token embeddings and the vocabulary output projection; we also report the total parameter count including both. Adjusting the expert count changes the router dimension, with a small effect on total computation that we include in the FLOPs accounting. Table~\ref{tab:budget_matching_accuracy} reports the maximum absolute budget deviations in percent. Actual configurations and deviations are given in Appendix~\ref{sec:budget_verification}.

\par

\begin{figure}[!t]
    \centering
    \begin{minipage}[t]{0.49\textwidth}
        \centering
        \includegraphics[height=0.405\linewidth]{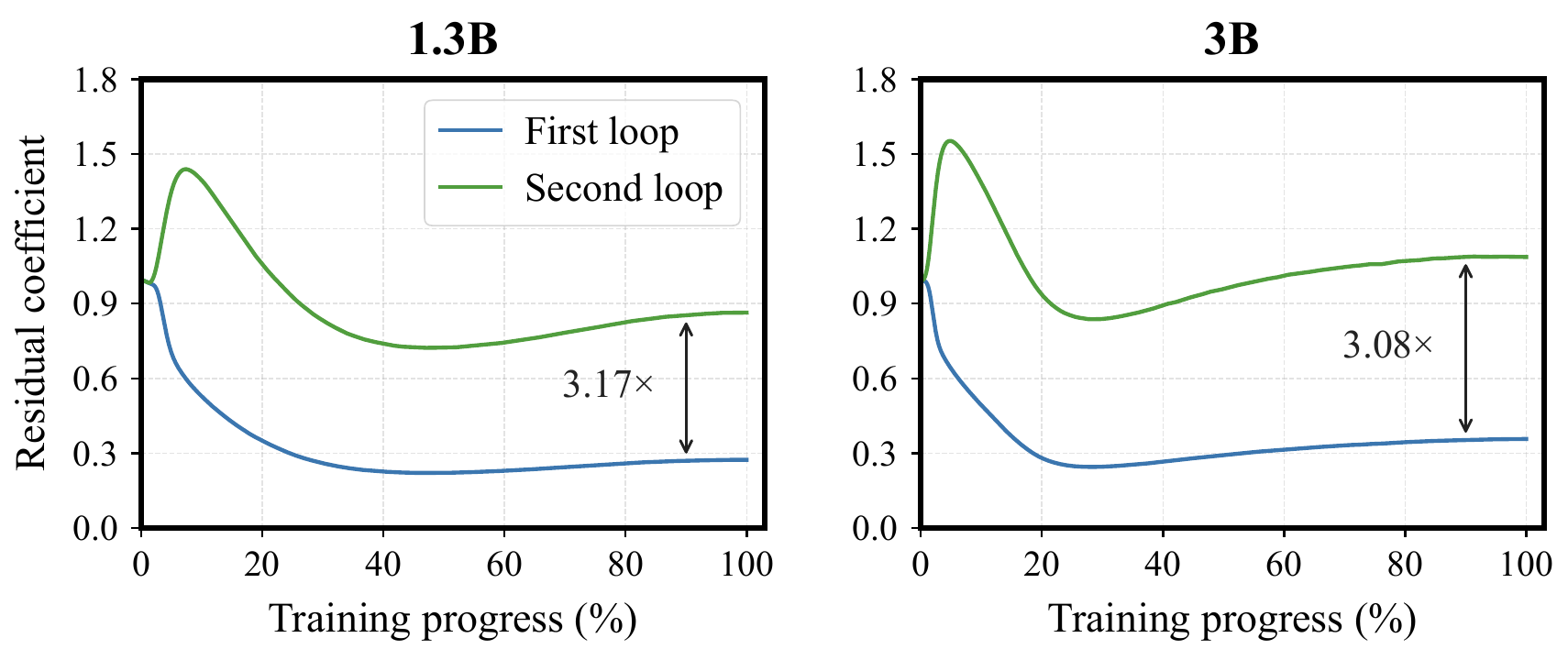}\\
        {\small (a) Learned residual coefficients}
    \end{minipage}\hfill
    \begin{minipage}[t]{0.49\textwidth}
        \centering
        \includegraphics[height=0.405\linewidth]{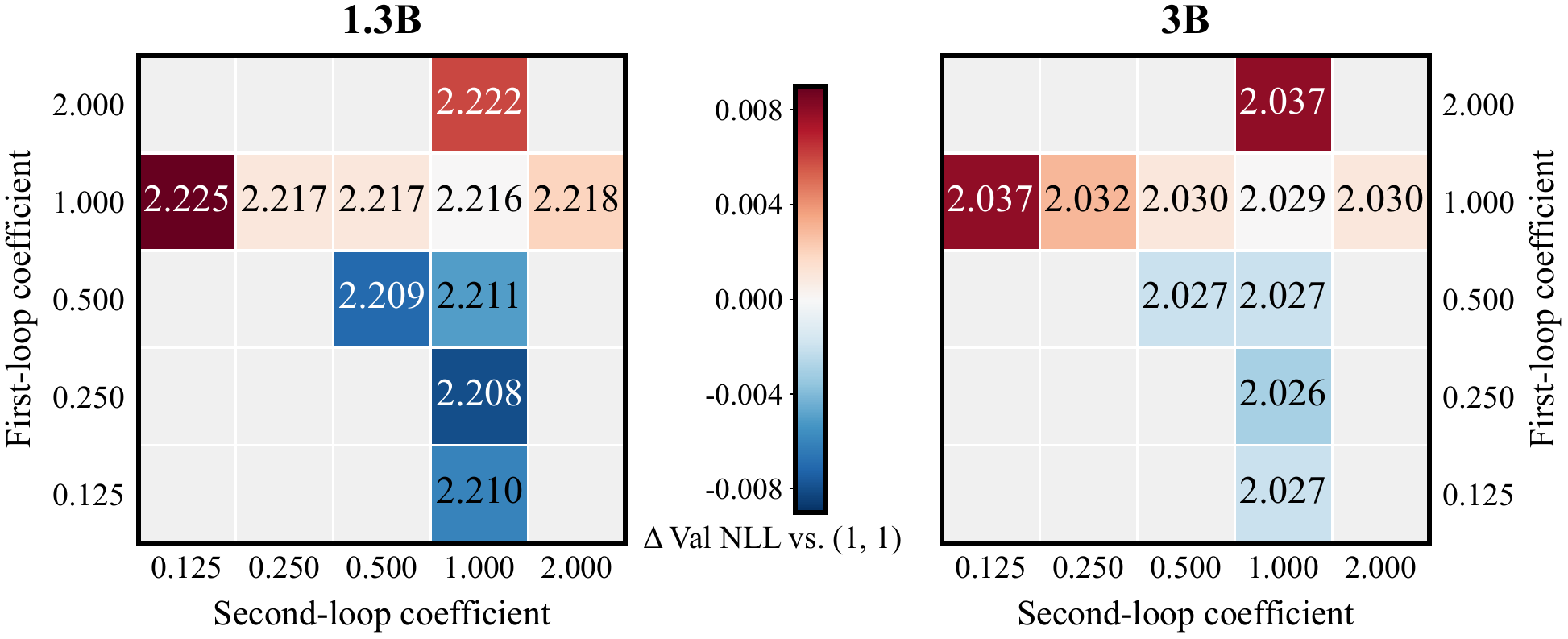}\\
        {\small (b) Fixed residual coefficients}
    \end{minipage}
    \caption{Residual scale asymmetry at 1.3B and 3B. (a) Learned coefficients for the two visits. (b) Validation NLL with fixed coefficients; rows and columns correspond to the first and second visits. Colors show differences from the same size $(1,1)$ baseline using scores rounded to three decimals.}
    \label{fig:learned_residuals}
    \label{fig:fixed_residuals}
\end{figure}

\section{LatentLoop}
\label{sec:latentloop}

\subsection{Asymmetry in Loop Updates}
\label{sec:asymmetry}

Although looped models can improve performance, the contribution of each visit remains incompletely understood. Prior work~\citep{geiping2026scaling,zhu2025scaling,wang2026smelt,blayney2026mechanistic} suggests that different loop visits can play different roles in processing and refining representations. We begin by asking whether the two visits contribute equally to the model's predictions. As an initial probe, we let each visit learn its own residual update scale and compare the resulting coefficients.

Specifically, we introduce two learnable scalar coefficients, $\alpha_1$ and $\alpha_2$, for the first and second visits. Within the repeated middle block, each residual branch is updated as
\begin{equation}
    h^{\prime} = h + \alpha_v f_{\ell}(h), \qquad v\in\{1,2\},
    \label{eq:visit_residual_scale}
\end{equation}
where $f_{\ell}$ denotes an attention or MoE residual branch. The same $\alpha_v$ is used for both branch types and all layers within visit $v$; entry and exit layers are unchanged. Both coefficients start at one and are optimized jointly with the model weights. We run this experiment on the 1.3B and 3B looped baselines at the 20B and 40B reference token budgets, respectively, using the same data and training settings as their fixed coefficient controls.

Figure~\ref{fig:learned_residuals}(a) shows a consistent asymmetry. During training, the first coefficient falls well below the second, while the second remains near one by the end. The final checkpoint coefficients are approximately $(0.27,0.87)$ for 1.3B and $(0.36,1.09)$ for 3B. Thus, joint optimization favors different update scales for the two visits at both model sizes. We also observe unequal residual scales with three and four visits at both model sizes, as shown in Section~\ref{sec:more_loop_analysis}.

Motivated by the learned asymmetry, we next ask how unequal residual scales affect performance. Since the larger learned coefficient remains close to one, we use one as a reference scale and vary which visit receives a different coefficient. Specifically, we fix the first or second coefficient at one and vary the other over $\{0.125,0.25,0.5,2\}$, training each setting from scratch. This compares the same scale change applied to different visits. Figure~\ref{fig:fixed_residuals}(b) shows that reducing the first coefficient does not hurt performance and even lowers validation NLL, whereas the corresponding reductions to the second coefficient increase it. Among the tested settings, $(0.25,1)$ gives the lowest NLL at both sizes, improving over $(1,1)$ by about 0.008 and 0.003 for 1.3B and 3B, respectively. Such a simple change to a single residual coefficient improves performance at both model sizes, without adding trainable parameters or changing the computational budget.

\citet{wang2026residual} propose symmetric residual scaling of $1/N$ across $N$ visits. In our two-visit experiments, the asymmetric setting $(0.25,1)$ achieves lower validation NLL than their corresponding $(0.5,0.5)$ setting at both model sizes (Figure~\ref{fig:fixed_residuals}(b)).

Together, the learned coefficients and fixed coefficient comparisons indicate a preference for asymmetric update scales in our looped MoE setting, motivating us to explore asymmetric compute allocation across visits.

\subsection{Asymmetric Compute Allocation}
\label{sec:allocation}

\begin{wrapfigure}{R}{0.48\textwidth}
    \centering
    \includegraphics[width=\linewidth,height=0.70\textheight,keepaspectratio]{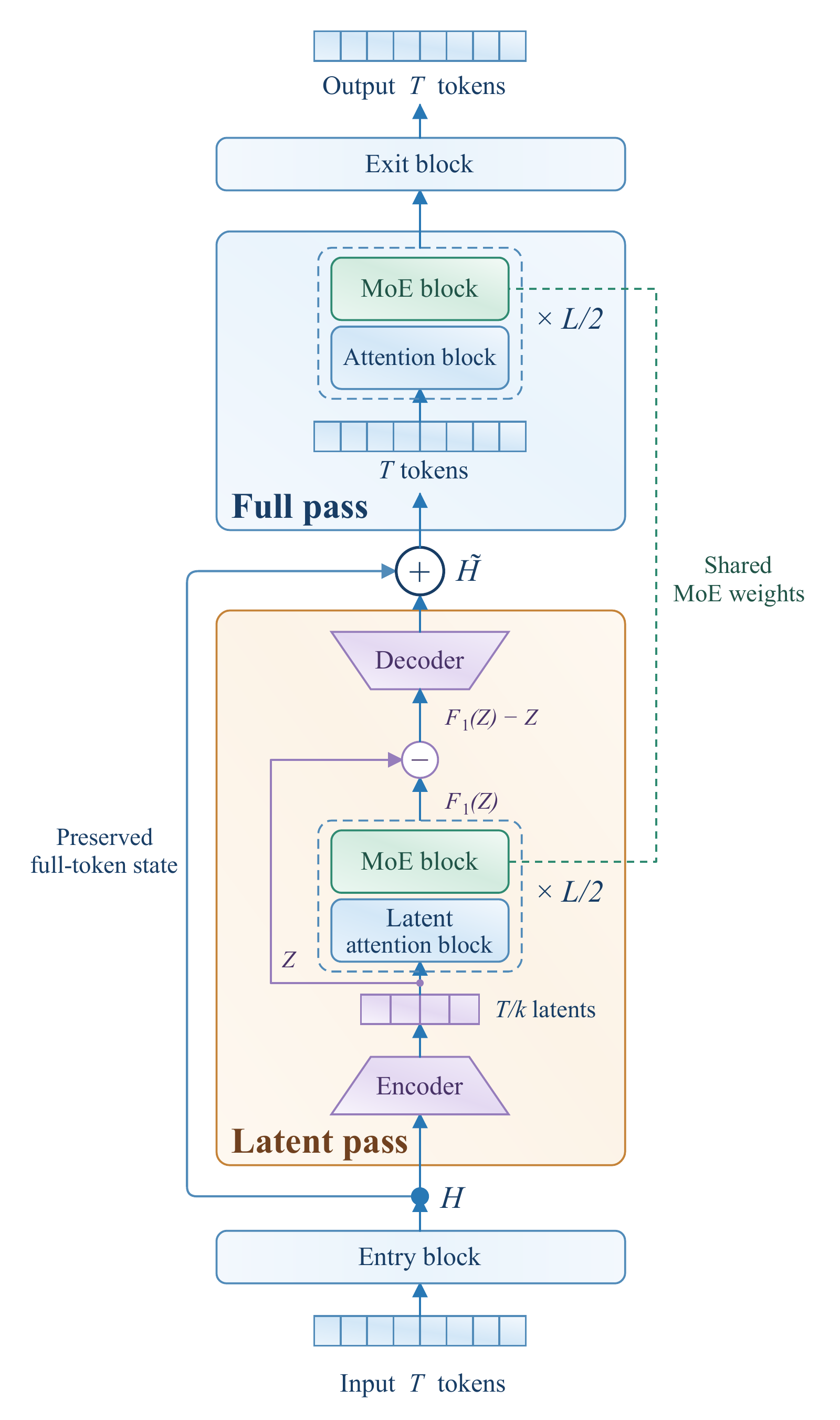}
    \caption{LatentLoop Architecture. LatentLoop compresses the first visit's sequence and retains full resolution in the second, sharing MoE weights with independent attention blocks.}
    \label{fig:latentloop_architecture}
\end{wrapfigure}

Reducing the first visit's residual coefficient can preserve or even improve performance. This observation motivates us to explore reducing computation during the first visit and reallocating the saved budget to other parts of the model.

We explore four ways to reduce first visit computation. First, inspired by projection into lower dimensional latent spaces~\citep{liu2024deepseek,elango2026latentmoe}, we project the residual stream to half its width, execute the first visit using nested submatrices of the shared weights, and project its increment back to the original width. Second, inspired by nested feed-forward networks~\citep{devvrit2024matformer}, we halve only the expert intermediate width using the corresponding submatrices, while keeping the residual stream and attention at full width. Third, we reduce the number of experts activated for each token from eight to four during the first visit. Fourth, inspired by sequence compression~\citep{hwang2026dynamic,yu2023megabyte,yu2026spiralformer}, we compress the sequence into fewer latent positions. These initial variants retain full parameter sharing and unit residual coefficients, with mean aggregation and direct writeback for sequence compression.

We find that sequence compression gives the lowest validation NLL among the four initial strategies and select it for further refinement. A possible explanation concerns how the strategies use the available parameter capacity. Looping saves stored parameters by reusing weights across visits. Under budget matching, MoE allows this saving to be allocated to a larger expert pool, preserving parameter capacity while matching computation. However, the other three reductions restrict the capacity available to each token during the first visit: activating fewer experts limits its expert capacity, while reducing widths uses only part of the shared weight matrices. Sequence compression instead retains the full model width and expert activation count at each latent position. Its saved computation is reallocated to model width, combining fewer computed positions with wider computation across the backbone. Results are reported in Section~\ref{sec:compute_reductions}.

Based on this comparison, we propose LatentLoop (Figure~\ref{fig:latentloop_architecture}). The first visit operates in a latent sequence space with fewer positions, while the second operates at the original sequence resolution. We reallocate the computation saved by the first visit to model width under the budget matching constraints in Section~\ref{sec:budget_matching}, creating an asymmetric allocation across visits.

Let $H\in\mathbb{R}^{T\times d}$ denote the states after the entry block, with sequence length $T$ and model width $d$. For an integer compression factor $k$ dividing $T$, the encoder $\operatorname{Enc}$ maps these states to a latent sequence of length $l=T/k$, while the decoder $\operatorname{Dec}$ maps a latent update back to $T$ positions. Both preserve the channel width $d$. The two visits follow
\begin{equation}
    \begin{aligned}
        Z &= \operatorname{Enc}(H), & Z&\in\mathbb{R}^{l\times d},\\
        \Delta &= F_1(Z)-Z,\\
        \widetilde H &= H+\operatorname{Dec}(\Delta),\\
        Y &= F_2(\widetilde H),
    \end{aligned}
    \label{eq:latentloop_flow}
\end{equation}
where $F_1$ and $F_2$ denote the first and second visits to the middle block. The original states $H$ are retained while the first visit processes $Z$; its increment $\Delta$ is then written back with causal alignment. The second visit processes the resulting full sequence $\widetilde H$, and its output $Y$ is passed to the exit block.

This asymmetric allocation also raises questions about update strength and parameter reuse. The first visit produces an increment in latent space that must be added to full resolution states. Together with the residual scale asymmetry in Section~\ref{sec:asymmetry}, this motivates learning how strongly the latent increment should be written back, as described in Section~\ref{sec:writeback}. Also, the two visits process different sequence resolutions, raising the question of whether direct reuse of the same block parameters remains appropriate for both. We investigate this choice in Section~\ref{sec:sharing}.

\subsection{Calibrated Writeback}
\label{sec:writeback}

We begin with a compression factor $k=2$ and use a simple mean encoder with direct writeback. Indexing the original and latent sequences from one, we aggregate each pair of adjacent states as
\begin{equation}
    Z_j=\operatorname{Enc}(H)_j=\frac{H_{2j-1}+H_{2j}}{2},
    \qquad 1\leq j\leq l.
    \label{eq:mean_encoder}
\end{equation}

After the first visit, direct writeback adds the latent increment $\Delta_j$ unchanged to positions $2j$ and $2j+1$, whenever these positions exist. The first receiver is the last token of the source pair; the second is the next token. This alignment ensures that a latent update only reaches positions where all tokens in its source pair are already available. The first token of the entire sequence, at position one, receives no latent update.

Mean aggregation and direct writeback fix the scales of both operators. The residual scaling observations in Section~\ref{sec:asymmetry} motivate us to let the model adjust these scales. We also explore a causal channel convolution encoder that learns the aggregation scale (Appendix~\ref{sec:additional_implementation}). On the decoder side, we introduce a learnable writeback coefficient for each channel and receiver position, which rescales the latent increment before it is added to the original states:
\begin{equation}
    \operatorname{Dec}(\Delta)_{kj+s}
    =w_s\odot\Delta_j,
    \qquad 0\leq s<k.
    \label{eq:dec2_writeback}
\end{equation}
Here $j=1,\ldots,l$, $kj+s\leq T$, and $\odot$ denotes elementwise multiplication. Positions before $k$ receive zero updates. Each vector $w_s\in\mathbb{R}^d$ starts at one and is shared across all token groups, so the decoder initially recovers direct writeback.

We compare learning the encoder, the decoder, or both. Learning either alone improves validation NLL over the initial design, while learning both provides no further improvement over the decoder alone in our comparison. Implementation details are provided in Appendix~\ref{sec:additional_implementation}, and ablation results in Section~\ref{sec:ablation_studies}.

We seek a simple and effective design with minimal added complexity. We therefore retain the parameter-free mean encoder and learn the writeback coefficients directly. With $k=2$, our \emph{scaled decoder} (SD) uses two vectors $w_0,w_1\in\mathbb{R}^d$ and adds only $2d$ parameters. It is applied once after the entire first visit and learns the update scale separately for each channel and receiver position.

\subsection{Selective Parameter Sharing}
\label{sec:sharing}

The first visit processes a compressed latent sequence, while the second processes the full sequence. Compression changes the sequence processed by attention and the hidden states passed to the MoE layers. The two resolutions may therefore favor different attention and MoE computations, motivating us to reconsider full parameter sharing across visits.

We approach this choice from the perspective of parameter allocation. MoE expert weights account for most of the stored backbone parameters, while attention occupies only a small fraction. We therefore give the two visits independent attention blocks within the repeated middle block, while retaining shared MoE routers and expert weights. In our matched 3B configuration, reducing the number of stored experts by just one per MoE layer, from 76 to 75, is sufficient to accommodate the additional attention block parameters within the parameter budget. Attention FLOPs remain unchanged, and the two visits already use separate KV caches, so independent attention blocks do not increase KV storage. Keeping the active expert count fixed also preserves expert FLOPs, with only a small change in routing cost. Untying the MoE instead would leave most parameters unshared across visits, making the model close to a non-looped Transformer in terms of parameter reuse. This allocation improves validation NLL over full parameter sharing under the budget constraints in Section~\ref{sec:budget_matching}. Detailed comparisons are reported in Section~\ref{sec:ablation_studies}.

\section{Experiments}
\label{sec:experiments}

\subsection{Experimental Setup}
\label{sec:experimental_setup}

\paragraph{Implementation details.}
We use sparse MoE Transformers in the style of Qwen3~\citep{yang2025qwen3}, with eight active experts per token. All methods share the same fixed FineWeb-Edu~\citep{penedo2024fineweb} training stream, TinyLlama tokenizer~\citep{zhang2024tinyllama}, and 0.1B token validation set. LatentLoop uses $k=2$, mean aggregation, and the scaled decoder, with shared MoE weights and independent attention blocks. Further implementation details are given in Appendix~\ref{sec:additional_implementation}.

\paragraph{Baselines.}
We compare with two baselines. \textbf{MoE Base} is a standard MoE Transformer with 64 stored experts per layer, executing each layer once. \textbf{MoE Loop} uses the looped architecture of SMELT~\citep{wang2026smelt} described in Section~\ref{sec:loop_architecture}, visiting the middle half of the layers twice while executing the entry and exit blocks once. Both visits share all middle block parameters and use residual coefficients of one. Detailed configurations and matched budgets are given in Appendix~\ref{sec:budget_verification}.

\paragraph{Benchmarks.}
We evaluate nine downstream benchmarks: LAMBADA OpenAI and LAMBADA Standard~\citep{paperno2016lambada}, HellaSwag~\citep{zellers2019hellaswag}, PIQA~\citep{bisk2020piqa}, WinoGrande~\citep{sakaguchi2021winogrande}, ARC-Easy and ARC-Challenge~\citep{clark2018think}, SciQ~\citep{welbl2017crowdsourcing}, and MMLU~\citep{hendrycks2020measuring} in its continuation formulation. We use \texttt{lm-evaluation-\allowbreak harness}~\citep{eval-harness} with both 0-shot and 5-shot settings for every benchmark. Additional details are given in Appendix~\ref{sec:additional_implementation}.

\begin{table}[!t]
    \centering
    \small
    \setlength{\tabcolsep}{3pt}
    \renewcommand{\arraystretch}{1.12}
    \resizebox{\textwidth}{!}{%
    \begin{tabular}{@{}cccl*{10}{c}@{}}
        \toprule
        \multicolumn{1}{@{}c}{\bfseries\boldmath Shot} & \multicolumn{1}{c}{\bfseries\boldmath Params} & \multicolumn{1}{c}{\bfseries\boldmath Token} & \multicolumn{1}{c}{\bfseries\boldmath Model} & \multicolumn{1}{c}{\bfseries\boldmath LD-O} & \multicolumn{1}{c}{\bfseries\boldmath LD-S} & \multicolumn{1}{c}{\bfseries\boldmath HS} & \multicolumn{1}{c}{\bfseries\boldmath PIQA} & \multicolumn{1}{c}{\bfseries\boldmath WG} & \multicolumn{1}{c}{\bfseries\boldmath ARC-E} & \multicolumn{1}{c}{\bfseries\boldmath ARC-C} & \multicolumn{1}{c}{\bfseries\boldmath SciQ} & \multicolumn{1}{c}{\bfseries\boldmath MMLU} & \multicolumn{1}{c@{}}{\bfseries\boldmath Avg} \\
        \midrule
         &  &  & MoE Base & 31.44 & 29.30 & 48.73 & 70.46 & 53.91 & 69.40 & 35.41 & 90.40 & 32.16 & 51.24 \\
         & 1.3B / 0.2B & 40B & MoE Loop & 31.15 & 28.72 & 49.61 & 69.91 & 56.04 & 67.42 & 36.26 & 89.60 & 31.86 & 51.18 \\
         &  &  & \cellcolor{gray!15}LatentLoop & \cellcolor{gray!15}32.99 & \cellcolor{gray!15}30.20 & \cellcolor{gray!15}50.36 & \cellcolor{gray!15}69.80 & \cellcolor{gray!15}56.35 & \cellcolor{gray!15}67.38 & \cellcolor{gray!15}37.12 & \cellcolor{gray!15}91.10 & \cellcolor{gray!15}31.93 & \cellcolor{gray!15}\textbf{51.91} \\
        \cmidrule(lr){2-14}
         &  &  & MoE Base & 40.89 & 36.17 & 57.89 & 73.18 & 57.93 & 73.40 & 42.49 & 93.10 & 35.12 & 56.69 \\
        5-shot & 3B / 0.5B & 70B & MoE Loop & 39.72 & 37.26 & 59.48 & 74.70 & 59.19 & 72.90 & 43.69 & 92.90 & 36.09 & 57.33 \\
         &  &  & \cellcolor{gray!15}LatentLoop & \cellcolor{gray!15}43.86 & \cellcolor{gray!15}41.74 & \cellcolor{gray!15}60.57 & \cellcolor{gray!15}74.65 & \cellcolor{gray!15}57.62 & \cellcolor{gray!15}74.41 & \cellcolor{gray!15}45.14 & \cellcolor{gray!15}92.80 & \cellcolor{gray!15}35.83 & \cellcolor{gray!15}\textbf{58.51} \\
        \cmidrule(lr){2-14}
         &  &  & MoE Base & 48.88 & 45.78 & 66.68 & 76.22 & 62.43 & 76.64 & 48.72 & 96.00 & 38.33 & 62.19 \\
         & 7B / 1B & 140B & MoE Loop & 50.07 & 43.66 & 68.69 & 75.95 & 64.09 & 76.18 & 49.32 & 95.80 & 39.22 & 62.55 \\
         &  &  & \cellcolor{gray!15}LatentLoop & \cellcolor{gray!15}49.21 & \cellcolor{gray!15}45.92 & \cellcolor{gray!15}68.44 & \cellcolor{gray!15}76.33 & \cellcolor{gray!15}63.93 & \cellcolor{gray!15}79.55 & \cellcolor{gray!15}51.71 & \cellcolor{gray!15}96.20 & \cellcolor{gray!15}39.35 & \cellcolor{gray!15}\textbf{63.40} \\
        \midrule
         &  &  & MoE Base & 39.78 & 31.92 & 48.99 & 69.75 & 53.51 & 61.15 & 32.17 & 79.10 & 30.74 & 49.68 \\
         & 1.3B / 0.2B & 40B & MoE Loop & 40.44 & 29.71 & 49.76 & 70.24 & 54.30 & 59.85 & 33.87 & 81.20 & 31.14 & 50.06 \\
         &  &  & \cellcolor{gray!15}LatentLoop & \cellcolor{gray!15}40.38 & \cellcolor{gray!15}31.98 & \cellcolor{gray!15}49.99 & \cellcolor{gray!15}69.86 & \cellcolor{gray!15}57.14 & \cellcolor{gray!15}60.44 & \cellcolor{gray!15}32.17 & \cellcolor{gray!15}81.20 & \cellcolor{gray!15}31.14 & \cellcolor{gray!15}\textbf{50.48} \\
        \cmidrule(lr){2-14}
         &  &  & MoE Base & 47.93 & 38.60 & 57.81 & 72.31 & 55.72 & 67.51 & 39.85 & 86.60 & 34.18 & 55.61 \\
        0-shot & 3B / 0.5B & 70B & MoE Loop & 47.70 & 39.65 & 59.38 & 73.83 & 57.62 & 67.21 & 38.57 & 85.60 & 34.62 & 56.02 \\
         &  &  & \cellcolor{gray!15}LatentLoop & \cellcolor{gray!15}49.23 & \cellcolor{gray!15}39.98 & \cellcolor{gray!15}60.26 & \cellcolor{gray!15}74.05 & \cellcolor{gray!15}56.51 & \cellcolor{gray!15}70.03 & \cellcolor{gray!15}39.93 & \cellcolor{gray!15}86.10 & \cellcolor{gray!15}34.15 & \cellcolor{gray!15}\textbf{56.69} \\
        \cmidrule(lr){2-14}
         &  &  & MoE Base & 54.12 & 45.80 & 65.83 & 76.17 & 59.83 & 69.82 & 44.54 & 89.10 & 36.45 & 60.18 \\
         & 7B / 1B & 140B & MoE Loop & 53.64 & 44.69 & 67.30 & 75.95 & 61.96 & 73.95 & 46.42 & 91.40 & 37.34 & 61.40 \\
         &  &  & \cellcolor{gray!15}LatentLoop & \cellcolor{gray!15}54.92 & \cellcolor{gray!15}45.22 & \cellcolor{gray!15}66.91 & \cellcolor{gray!15}75.68 & \cellcolor{gray!15}62.59 & \cellcolor{gray!15}74.20 & \cellcolor{gray!15}46.50 & \cellcolor{gray!15}90.40 & \cellcolor{gray!15}37.69 & \cellcolor{gray!15}\textbf{61.57} \\
        \bottomrule
    \end{tabular}%
    }
    \caption{Downstream benchmark scores (\%). Params denote approximate total/active sizes; token budgets are FLOPs matching reference labels. LD-O/LD-S denote LAMBADA OpenAI/Standard, HS is HellaSwag, WG is WinoGrande, and MMLU uses continuation scoring.}
    \label{tab:downstream}
\end{table}

\subsection{Scaling Analysis}
\label{sec:scaling_analysis}

\paragraph{Setup.}
\label{sec:scaling_setup}
We evaluate 15 matched budget configurations spanning 1.3B--7B parameters, with the largest training budget exceeding $10^{21}$ FLOPs. Budgets within each architecture and model size share a training mainline with separate annealing branches (Appendix~\ref{sec:additional_implementation}). Similar to SMELT~\citep{wang2026smelt}, we fit a separate additive power-law loss surface for each architecture:
\begin{equation}
    \widehat L(F,D)
    = E + A\left(\frac{F}{10^9}\right)^{-\alpha}
        + K\left(\frac{D}{10^{10}}\right)^{-\beta},
    \label{eq:scaling_surface}
\end{equation}
\begin{wrapfigure}{R}{0.48\textwidth}
    \centering
    \includegraphics[width=\linewidth]{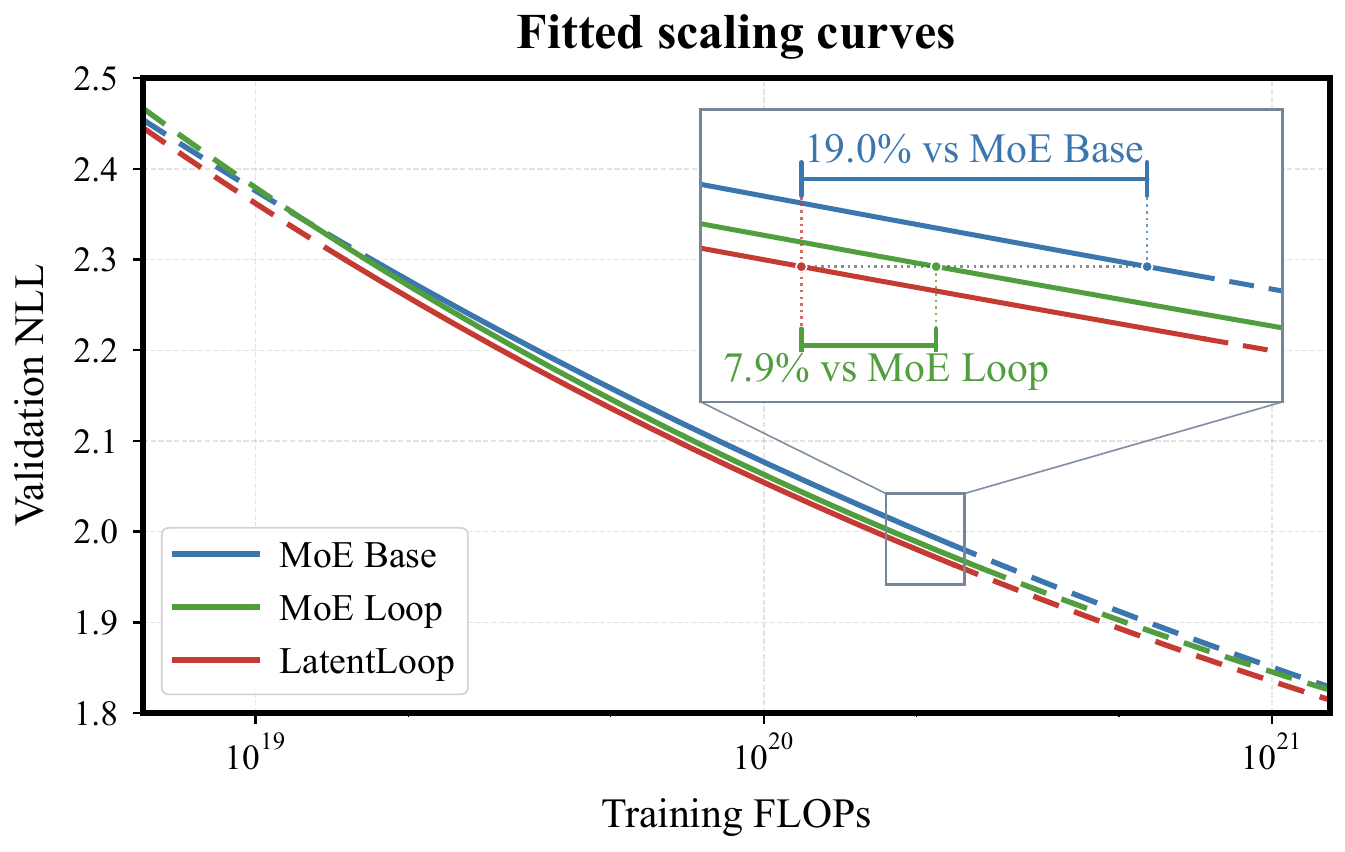}
    \caption{Fitted compute-optimal NLL. Solid segments lie within the convex hull of observed $(\log F,\log D)$ points; dashed segments involve extrapolation.}
    \label{fig:scaling_frontier}
    \end{wrapfigure}

where $F$ is theoretical training FLOPs per prediction token and $D$ is the cumulative number of training prediction tokens. We use the actual computation and token count of each run, including its shared training prefix and final annealing phase. We fit the five parameters with equal weight for each endpoint, minimizing Huber loss on the difference between predicted and observed log NLL, with $\delta=10^{-3}$ and 48 initializations of L-BFGS-B. For a total compute budget $C=FD$, we minimize the fitted loss over $F$ and $D$ to obtain the predicted compute-optimal frontier.

\paragraph{Results.}
\label{sec:scaling_results}
At all 15 matched budget endpoints, LatentLoop consistently achieves lower validation NLL than MoE Base and MoE Loop (Appendix~\ref{sec:full_scaling_results}). Figure~\ref{fig:scaling_frontier} summarizes this improvement through the fitted frontiers. Within the observed model and data allocation range, the fitted frontiers show that LatentLoop reaches an NLL of 1.988 with $1.85\times10^{20}$ FLOPs, saving 19.0\% of computation relative to MoE Base and 7.9\% relative to MoE Loop at the same loss. Together, the matched endpoint comparisons and fitted frontiers show that asymmetric compute allocation is an effective way to improve the compute efficiency of looped MoE models.

\subsection{Downstream Evaluation}
\label{sec:downstream_evaluation}

Table~\ref{tab:downstream} reports downstream scores at the largest reference token budget for each size: 40B, 70B, and 140B for the 1.3B, 3B, and 7B tiers, respectively. LatentLoop achieves the highest average score in all six size and shot settings. Relative to MoE Loop, the average gains range from 0.73 to 1.18 percentage points in 5-shot and from 0.17 to 0.67 in 0-shot. Relative to MoE Base, the average gains range from 0.67 to 1.82 percentage points in 5-shot and from 0.80 to 1.39 in 0-shot. These results show that the validation NLL improvements also carry over to average downstream performance, with larger gains in 5-shot.

\subsection{Ablation Studies}
\label{sec:ablation_studies}

\begin{wraptable}{R}{0.48\textwidth}
    \centering
    \small
\renewcommand{\arraystretch}{1.12}
\setlength{\tabcolsep}{3pt}
\begin{tabularx}{\linewidth}{@{}Xr@{}}
    \toprule
    {\bfseries\boldmath Strategy} & {\bfseries\boldmath NLL $\downarrow$} \\
    \midrule
    \multicolumn{2}{@{}l}{\textit{Baselines}} \\
    MoE Base & 2.0426 \\
    MoE Loop & 2.0293 \\
    \midrule
    \multicolumn{2}{@{}l}{\textit{Initial first visit reductions}} \\
    Residual width  & 2.0361 \\
    Expert intermediate width  & 2.0364 \\
    Active experts & 2.0349 \\
    Sequence length & 2.0303 \\
    \midrule
    \multicolumn{2}{@{}l}{\textit{Final design}} \\
    \rowcolor{gray!15}
    LatentLoop (ours) & \textbf{2.0199} \\
    \bottomrule
\end{tabularx}
\caption{Compute reduction strategies.}
\label{tab:compute_reductions}

\end{wraptable}

Unless otherwise stated, the following ablation and analysis experiments use the 3B/40B reference setting. We study compute allocation, the encoder, the decoder, and parameter sharing. Each ablation is trained from scratch with the training settings in Section~\ref{sec:experimental_setup}, following the budget matching procedure in Section~\ref{sec:budget_matching}.

\paragraph{Comparing compute reduction strategies.}
\label{sec:compute_reductions}
Table~\ref{tab:compute_reductions} compares the four initial strategies described in Section~\ref{sec:allocation}, with fully shared block parameters and unit residual coefficients. Sequence aggregation with direct writeback gives the lowest NLL among these strategies. Sequence compression preserves full width and expert activation at each latent position, which may explain its advantage over reducing widths or active expert counts. Based on these initial results, we select sequence aggregation for further refinement. The encoder, decoder, and parameter sharing comparisons below yield the final LatentLoop design, reported separately at 2.0199.

\begin{table}[!htbp]
    \centering
    \begin{minipage}[t]{0.48\textwidth}
    \vspace{0pt}
        \centering
        \small
\renewcommand{\arraystretch}{1.12}
\setlength{\tabcolsep}{3pt}
\begin{tabularx}{\linewidth}{@{}XXr@{}}
    \toprule
    {\bfseries\boldmath Encoder} & {\bfseries\boldmath Decoder} & {\bfseries\boldmath NLL $\downarrow$} \\
    \midrule
    \multicolumn{3}{@{}l}{\textit{Baselines}} \\
    \multicolumn{2}{@{}l}{MoE Base} & 2.0426 \\
    \multicolumn{2}{@{}l}{MoE Loop} & 2.0293 \\
    \midrule
    \multicolumn{3}{@{}l}{\textit{SpiralFormer style operators}} \\
    Mean & Direct & 2.0268 \\
    Softmax & Softmax & 2.0264 \\
    \midrule
    \multicolumn{3}{@{}l}{\textit{Our enc/dec variants}} \\
    Mean & Direct & 2.0268 \\
    Conv & Direct & 2.0219 \\
    \rowcolor{gray!15}
    Mean & SD (ours) & \textbf{2.0199} \\
    Conv & SD & 2.0209 \\
    \midrule
    \multicolumn{3}{@{}l}{\textit{Writeback scale controls}} \\
    Mean & SD (RMS) & 2.0228 \\
    Mean & Scalar gain & 2.0210 \\
    \bottomrule
\end{tabularx}
\caption{Encoder and decoder comparisons. SD denotes the scaled decoder. All operator variants use independent attention blocks across visits.}
\label{tab:ablation_encdec}

    \end{minipage}\hfill
    \begin{minipage}[t]{0.48\textwidth}
    \vspace{0pt}
        \centering
        \small
\renewcommand{\arraystretch}{1.12}
\setlength{\tabcolsep}{3pt}
\begin{tabularx}{\linewidth}{@{}Xcr@{}}
    \toprule
    {\bfseries\boldmath Model / operators} & {\bfseries\boldmath Attn block} & {\bfseries\boldmath NLL $\downarrow$} \\
    \midrule
    \multicolumn{3}{@{}l}{\textit{Baselines}} \\
    MoE Base & -- & 2.0426 \\
    MoE Loop & Shared & 2.0293 \\
    \midrule
    \multicolumn{3}{@{}l}{\textit{MoE Loop variants}} \\
    MoE Loop & Shared & 2.0293 \\
    MoE Loop & Untied & 2.0319 \\
    \midrule
    \multicolumn{3}{@{}l}{\textit{Our variants: Mean + Direct}} \\
    Mean + Direct & Shared & 2.0303 \\
    Mean + Direct & Untied & 2.0268 \\
    \midrule
    \multicolumn{3}{@{}l}{\textit{Our variants: Mean + SD}} \\
    Mean + SD & Shared & 2.0237 \\
    \rowcolor{gray!15}
    Mean + SD (ours) & Untied & \textbf{2.0199} \\
    \bottomrule
\end{tabularx}
\caption{Parameter sharing comparisons.}
\label{tab:ablation_sharing}

    \end{minipage}
\end{table}

\paragraph{Encoder and decoder.}

Table~\ref{tab:ablation_encdec} compares the encoding and writeback operators with independent attention blocks across visits. Direct denotes direct writeback, and Conv denotes channel convolution. We include softmax aggregation and allocation inspired by SpiralFormer~\citep{yu2026spiralformer}. These operators give an NLL of 2.0264, slightly below mean aggregation and direct writeback.

Relative to mean aggregation and direct writeback at 2.0268, learning channel convolution coefficients on the encoder side lowers NLL to 2.0219, while learning the writeback coefficients of the scaled decoder (SD) with the mean encoder lowers it to 2.0199. Learning both together gives 2.0209, providing no further reduction over the scaled decoder alone in this comparison. We therefore retain the mean encoder and scaled decoder for a simple design with only $2d$ added parameters.

To separate overall writeback scale from relative channel and receiver position adjustments, we add two controls based on mean aggregation and the scaled decoder. In the normalized variant, we jointly divide the $2d$ writeback coefficients by their RMS and use no additional learned gain. This retains relative coefficients but fixes their global RMS to one. In the scale only variant, a single learned scalar replaces all channel and position coefficients. The two variants give NLLs of 2.0228 and 2.0210, respectively, compared with 2.0268 for direct writeback and 2.0199 for the scaled decoder. A single scalar therefore recovers most of the observed improvement, while channel and position adjustments provide a smaller additional reduction. The higher NLL under coefficient normalization also suggests that allowing the overall scale to adapt matters in this setting.
\par

\paragraph{Parameter sharing.}

Table~\ref{tab:ablation_sharing} compares parameter sharing with the stored expert count adjusted to match the parameter budget. For ordinary MoE Loop, with two full resolution visits, reducing the expert count from 92 to 91 while untying attention blocks increases NLL from 2.0293 to 2.0319. In contrast, the corresponding change from 76 to 75 experts improves both compressed first visit variants. NLL falls from 2.0303 to 2.0268 with mean aggregation and direct writeback, and from 2.0237 to 2.0199 with mean aggregation and the scaled decoder. Thus, in our matched experiments, independent attention blocks benefit the two latent sequence configurations, but not ordinary MoE Loop. These results support our choice to share MoE routers and experts while allowing separate attention blocks across the two resolutions.

Concurrent with our work, \citet{wang2026loop} report improvements from untying attention in looped MoE models. However, their tied versus untied comparisons do not reduce the expert count to compensate for the additional attention parameters, so the untied models have more total parameters. Our comparison instead accounts for these additional parameters by reducing the expert count. This difference may contribute to the contrasting observation for ordinary MoE Loop.

Additional ablations are reported in Appendix~\ref{sec:additional_ablations}.

\subsection{Analysis}
\label{sec:analysis}

\paragraph{Residual scales with more loop visits.}
\label{sec:more_loop_analysis}

The two visits in Section~\ref{sec:asymmetry} learn different residual scales, with a smaller coefficient for the first visit. We next ask whether this asymmetry persists when the model makes more visits. We extend the experiment to three and four visits of the shared middle half at both 1.3B and 3B. Each visit has its own learnable residual coefficient, initialized at one and shared across its attention and MoE branches. We retain the 20B and 40B reference token budgets for 1.3B and 3B, respectively, and apply the budget matching procedure in Section~\ref{sec:budget_matching}.

\begin{figure}[!htbp]
    \centering
    \includegraphics[width=\textwidth]{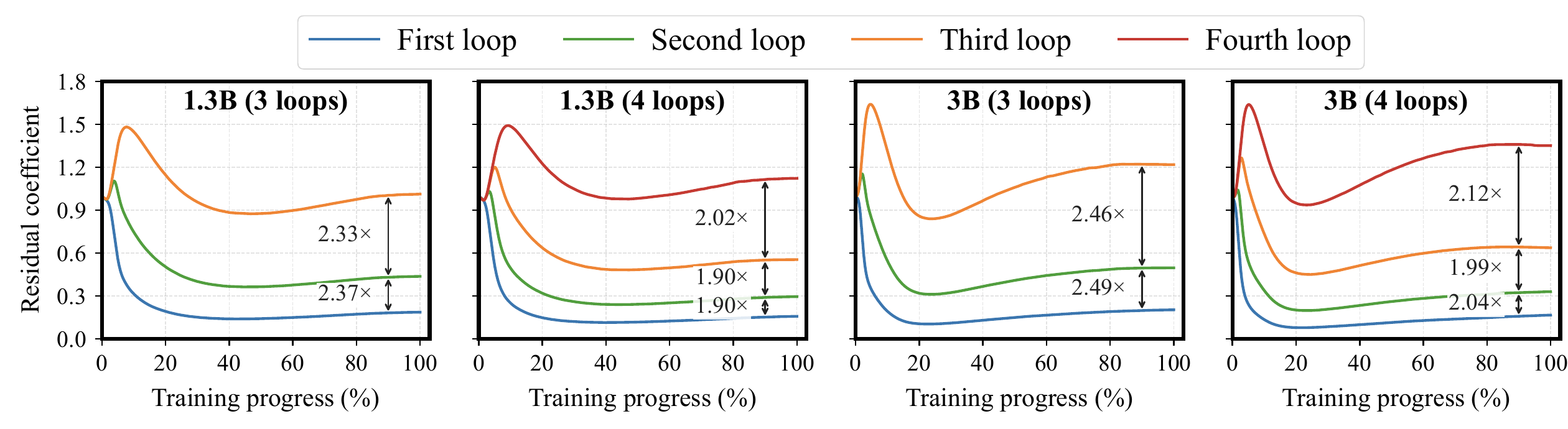}
    \caption{Learned residual coefficients with three and four visits, for 1.3B and 3B.}
    \label{fig:more_loop_residuals}
\end{figure}

Figure~\ref{fig:more_loop_residuals} shows a consistent separation of the coefficients. Near the end of training, the coefficients increase with visit index in all four experiments: the first visit has the smallest scale, and the last has the largest. Thus, the preference for unequal update scales also appears beyond two visits. With more visits, we additionally observe similar ratios between consecutive coefficients within each run. At 90\% training progress, the ratios marked in the figure are approximately $2.4\times$ for three visits and $2.0\times$ for four visits. We leave explaining this pattern to future work.

\paragraph{Learned writeback coefficients.}
\label{sec:further_analysis}

\begin{wrapfigure}{R}{0.48\textwidth}
    \centering
    \includegraphics[width=\linewidth]{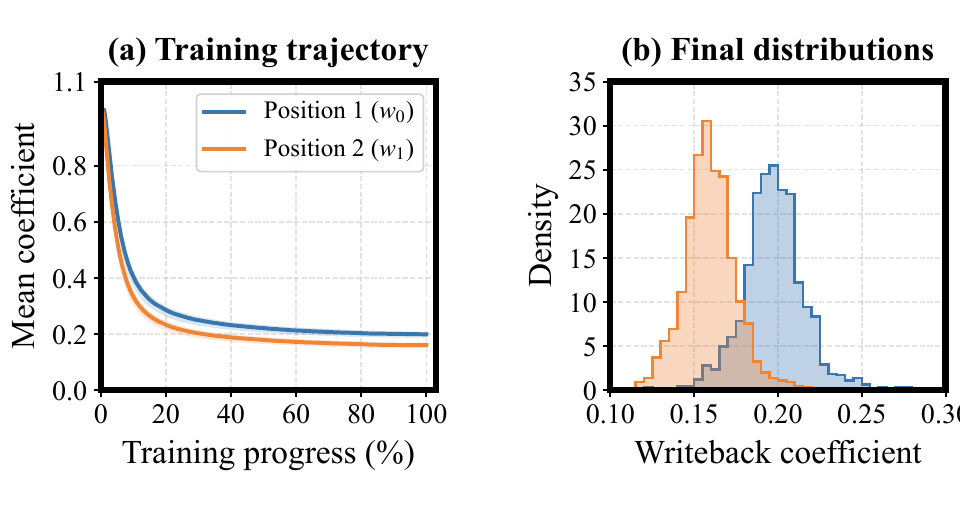}
    \caption{Learned scaled decoder coefficients at 3B/40B. (a) Channel means during training, with 25th--75th percentile bands. (b) Final channel distributions.}
    \label{fig:dec2_coefficients}
\end{wrapfigure}

Figure~\ref{fig:dec2_coefficients} shows that both receiver positions learn scales well below their unit initialization, with final channel means of approximately 0.199 and 0.160. This echoes the preference for a smaller first visit residual scale in Section~\ref{sec:asymmetry}. After warmup, the nearer receiver, $2j$, consistently learns a higher mean coefficient than the farther receiver, $2j+1$. This pattern is consistent with the latent update providing more reliable information for the nearer position: the source pair includes the state at $2j$, whereas the update received at $2j+1$ comes from the preceding pair and does not include its current state. The scaled decoder thus learns to give greater weight to the update at the nearer receiver.

\section{Related Work}
\label{sec:related_work}

\paragraph{Mixture-of-experts Transformers.}
Mixture-of-experts (MoE) models distribute computation among multiple expert networks through a learned gating mechanism~\citep{shazeer2017outrageously}. We use sparse MoE Transformers~\citep{fedus2022switch,liu2024deepseek} in the style of Qwen3~\citep{yang2025qwen3}, which activate only a small subset of experts for each token, allowing parameter capacity to grow without a proportional increase in computation per token. Recent work~\citep{jaggi2026tying,qiu2026more,tan2025rexmoe,huang2026unipool,chen2026mixture} in this area explores reusing experts across depth through sharing within neighboring layers, access to experts from other layers, or a common expert pool. Sharing expert weights while keeping attention independent is a design choice that we also adopt. For example, MoRE~\citep{qiu2026more} lets two adjacent layers draw from a larger shared expert pool while retaining separate attention and routers. Unlike MoRE, LatentLoop revisits an ordered stack of middle layers: each visit traverses the same sequence of expert pools, with shared routers and experts but independent attention blocks.

\paragraph{Looped Transformers.}
Recurrent models separate computational depth from the number of stored parameters. Universal Transformers~\citep{dehghani2018universal} apply a shared block repeatedly with adaptive halting, while deep equilibrium models~\citep{bai2019deep} represent repeated computation through an implicit fixed point. More recently, Huginn~\citep{geiping2026scaling} and Ouro~\citep{zhu2025scaling} scale recurrent computation and demonstrate reasoning gains from additional loop visits. These approaches and related studies~\citep{prairie2026parcae,wang2026residual,blayney2026mechanistic} focus mainly on dense models. However, comparisons at fixed parameter counts can give looped models more computation, while matching computation alone can leave them with fewer stored parameters. This motivates looped MoE models~\citep{wang2026smelt,chen2026loopmoe} that jointly match parameter and compute budgets.

\paragraph{Looped MoE Transformers.}
MoE offers a way to balance these budgets because the number of stored experts can be adjusted separately from the number activated per token. \citet{lee2026sparse} find that sparse layers improve the scaling of looped models and associate this gain with routing divergence across visits. LoopMoE~\citep{chen2026loopmoe} combines iteration conditioned normalization with capacity balancing to compare against a non-looped MoE at matched total parameters and per-token FLOPs. \citet{chen2026scaling} jointly model recurrence and sparsity, finding diminishing gains from additional loops that depend on the degree of sparsity. SMELT~\citep{wang2026smelt} matches per-token FLOPs, stored backbone parameters, and KV cache, and finds that looping the middle half of layers twice improves validation loss and downstream performance. Its analysis shows that the second visit reinforces earlier residual updates and reduces attention sinks. Building on SMELT's budget matching setting, we study how to improve the allocation within a two-visit loop. Our residual coefficient observations motivate different compute allocations across visits, leading to compressed first-visit computation, calibrated writeback, and selective parameter sharing.

\paragraph{Compressed computation.}
Prior work~\citep{yu2023megabyte,hwang2026dynamic,barrault2024large} explores computation over compressed or multiscale representations, using byte patches, learned chunks, or sentence embeddings. Most closely related to ours, SpiralFormer~\citep{yu2026spiralformer} also applies sequence compression within looped Transformers. Its main focus is learning hierarchical dependencies through a schedule of multiple sequence resolutions. In contrast, our work builds on the MoE architecture and focuses on fair comparisons under matched budgets. We use compression to allocate computation asymmetrically across visits and reallocate the saved computation to model width. We further study learned writeback calibration and selective parameter sharing across visits.

In our experiments, we directly compare SpiralFormer-style mean and softmax aggregation and writeback operators (Table~\ref{tab:ablation_encdec}) and show that our design achieves lower validation NLL. We also compare our largest LatentLoop model against the largest SpiralFormer-L, with details in Appendix~\ref{sec:spiralformer_comparison}.

\section{Conclusion}
\label{sec:conclusion}

We studied how to improve looped MoE models through compute allocation under matched resource budgets. Learned and fixed residual coefficient experiments reveal an asymmetry in the preferred update scales of the two visits, motivating us to explore different amounts of computation across visits. LatentLoop processes a compressed sequence during the first visit and the full sequence during the second, reallocating the saved computation to model width. Learned writeback coefficients calibrate the latent increment, while shared MoE routers and experts preserve parameter reuse alongside independent attention blocks.
Across 15 budget configurations at the 1.3B, 3B, and 7B tiers, with the largest training budget exceeding $10^{21}$ FLOPs, LatentLoop consistently achieves lower validation NLL than MoE Base and MoE Loop. It also obtains higher average scores across nine downstream benchmarks at all three sizes in both 0-shot and 5-shot settings. Our experiments suggest that asymmetric compute allocation is a promising direction for improving looped MoE models.

\label{sec:main_end}

\section*{Limitations}
\label{sec:limitations}
LatentLoop's asymmetric compute allocation is motivated by the observed asymmetry in residual coefficients. However, the mechanisms underlying this learned asymmetry remain incompletely understood, particularly the residual coefficient patterns observed with three and four loop visits in Section~\ref{sec:more_loop_analysis}. We leave a more detailed investigation of these patterns to future work. Further study of the specific mechanisms and roles of different loop visits may guide more effective compute allocation for looped MoE models.

\begingroup
\setlength{\bibsep}{3pt plus 1pt minus 1pt}
\bibliographystyle{plainnat}
\bibliography{references}
\endgroup

\clearpage
\appendix
\section{Model Configurations and Matched Budgets}
\label{sec:budget_verification}

Table~\ref{tab:actual_architectures} lists the nine architectures used for the three methods at the three parameter tiers. Each architecture is unchanged across its token budgets. All configurations use eight query heads, two KV heads, eight active experts per token, context length 4096, and global batch size 256. The resulting widths, head dimensions, and expert counts may be less favorable for hardware execution; these values are chosen to match the resource budgets for fair comparisons. $P_b$ counts the backbone, including attention, experts, routers, normalization, and any learned decoder. It excludes the token embedding and vocabulary output projection; $P_{\rm all}$ includes both. Shared parameters are counted once, and the independent attention block parameters in LatentLoop are included.


\begin{table}[!htbp]
    \centering
    \begin{minipage}{0.88\linewidth}
    \centering
    \small
    \setlength{\tabcolsep}{2pt}
    \renewcommand{\arraystretch}{1.12}
    \begin{tabularx}{\linewidth}{@{}Xcrrrrrr@{}}
        \toprule
        \multicolumn{1}{@{}c}{\bfseries\boldmath Model} & \multicolumn{1}{c}{\bfseries\boldmath Layers} & \multicolumn{1}{c}{\bfseries\boldmath $d$} & \multicolumn{1}{c}{\bfseries\boldmath $d_e$} & \multicolumn{1}{c}{\bfseries\boldmath $d_h$} & \multicolumn{1}{c}{\bfseries\boldmath $E$} & \multicolumn{1}{c}{\bfseries\boldmath $P_b$} & \multicolumn{1}{c@{}}{\bfseries\boldmath $P_{\rm all}$} \\
        \midrule
        \multicolumn{8}{c}{\textbf{1.3B}} \\
        MoE Base & 12/12 & 1024 & 512 & 128 & 64 & 1.240232 & 1.305768 \\
        MoE Loop & 12/18 & 864 & 432 & 86 & 91 & 1.241561 & 1.296857 \\
        \rowcolor{gray!15}
        LatentLoop & 12/18 & 944 & 472 & 102 & 75 & 1.238587 & 1.299003 \\
        \midrule
        \multicolumn{8}{c}{\textbf{3B}} \\
        MoE Base & 16/16 & 1408 & 704 & 176 & 64 & 3.125857 & 3.215969 \\
        MoE Loop & 16/24 & 1184 & 592 & 118 & 92 & 3.141792 & 3.217568 \\
        \rowcolor{gray!15}
        LatentLoop & 16/24 & 1296 & 648 & 140 & 75 & 3.112028 & 3.194972 \\
        \midrule
        \multicolumn{8}{c}{\textbf{7B}} \\
        MoE Base & 20/20 & 1856 & 928 & 232 & 64 & 6.788591 & 6.907375 \\
        MoE Loop & 20/30 & 1552 & 776 & 154 & 93 & 6.818844 & 6.918172 \\
        \rowcolor{gray!15}
        LatentLoop & 20/30 & 1696 & 848 & 186 & 76 & 6.750198 & 6.858742 \\
        \bottomrule
    \end{tabularx}
    \caption{Actual architecture configurations. The Layers column gives stored/executed depth; $d$, $d_e$, and $d_h$ are model width, expert intermediate width, and attention head dimension. $E$ is the stored expert count per layer. Parameter counts are in billions.}
    \label{tab:actual_architectures}
    \end{minipage}
\end{table}

\begin{table}[!htbp]
    \centering
    \begin{minipage}{0.88\linewidth}
    \centering
    \small
    \setlength{\tabcolsep}{1.4pt}
    \renewcommand{\arraystretch}{1.12}
    \begin{tabularx}{\linewidth}{@{}Xrrrrrr@{}}
        \toprule
        \multicolumn{1}{@{}c}{\bfseries\boldmath Model} & \multicolumn{1}{c}{\bfseries\boldmath KV (MiB)} & \multicolumn{1}{c}{\bfseries\boldmath $F_{\rm step}$} & \multicolumn{1}{c}{\bfseries\boldmath $\Delta P_b$} & \multicolumn{1}{c}{\bfseries\boldmath $\Delta P_{\rm all}$} & \multicolumn{1}{c}{\bfseries\boldmath $\Delta$KV} & \multicolumn{1}{c@{}}{\bfseries\boldmath $\Delta F_{\rm step}$} \\
        \midrule
        \multicolumn{7}{c}{\textbf{1.3B}} \\
        MoE Base & 48.00 & 1.676 & +0.00 & +0.00 & +0.00 & +0.00 \\
        MoE Loop & 48.38 & 1.685 & +0.11 & -0.68 & +0.78 & +0.54 \\
        \rowcolor{gray!15}
        LatentLoop & 47.81 & 1.672 & -0.13 & -0.52 & -0.39 & -0.25 \\
        \midrule
        \multicolumn{7}{c}{\textbf{3B}} \\
        MoE Base & 88.00 & 3.767 & +0.00 & +0.00 & +0.00 & +0.00 \\
        MoE Loop & 88.50 & 3.801 & +0.51 & +0.05 & +0.57 & +0.90 \\
        \rowcolor{gray!15}
        LatentLoop & 87.50 & 3.786 & -0.44 & -0.65 & -0.57 & +0.50 \\
        \midrule
        \multicolumn{7}{c}{\textbf{7B}} \\
        MoE Base & 145.00 & 7.630 & +0.00 & +0.00 & +0.00 & +0.00 \\
        MoE Loop & 144.38 & 7.650 & +0.45 & +0.16 & -0.43 & +0.26 \\
        \rowcolor{gray!15}
        LatentLoop & 145.31 & 7.646 & -0.57 & -0.70 & +0.22 & +0.21 \\
        \bottomrule
    \end{tabularx}
    \caption{Single-step resources and deviations from MoE Base at the same tier. KV is BF16 K/V tensor storage per 4096-token sequence; $F_{\rm step}$ is in PFLOPs. All $\Delta$ columns are signed percentage differences computed before rounding.}
    \label{tab:resource_deviations}
    \end{minipage}
\end{table}

\begin{table}[!htbp]
    \centering
    \begin{minipage}{0.88\linewidth}
    \centering
    \small
    \setlength{\tabcolsep}{2pt}
    \renewcommand{\arraystretch}{1.12}
    \begin{tabularx}{\linewidth}{@{}Xrrrrr@{}}
        \toprule
        \multicolumn{1}{@{}c}{\bfseries\boldmath Model} & \multicolumn{1}{c}{\bfseries\boldmath Updates} & \multicolumn{1}{c}{\bfseries\boldmath Token} & \multicolumn{1}{c}{\bfseries\boldmath $\Delta D$} & \multicolumn{1}{c}{\bfseries\boldmath $C$} & \multicolumn{1}{c@{}}{\bfseries\boldmath $\Delta C$} \\
        \midrule
        \multicolumn{6}{c}{\textbf{1.3B}} \\
        MoE Base & 38,146 & 39.999B & +0.00 & 0.6392 & +0.00 \\
        MoE Loop & 37,941 & 39.784B & -0.54 & 0.6392 & +12.47 \\
        \rowcolor{gray!15}
        LatentLoop & 38,240 & 40.098B & +0.25 & 0.6392 & -0.98 \\
        \midrule
        \multicolumn{6}{c}{\textbf{3B}} \\
        MoE Base & 66,757 & 70.000B & +0.00 & 2.5146 & +0.00 \\
        MoE Loop & 66,161 & 69.375B & -0.89 & 2.5146 & +0.02 \\
        \rowcolor{gray!15}
        LatentLoop & 66,427 & 69.654B & -0.49 & 2.5146 & -3.57 \\
        \midrule
        \multicolumn{6}{c}{\textbf{7B}} \\
        MoE Base & 133,514 & 140.000B & +0.00 & 10.1877 & +0.00 \\
        MoE Loop & 133,170 & 139.639B & -0.26 & 10.1877 & +2.27 \\
        \rowcolor{gray!15}
        LatentLoop & 133,240 & 139.712B & -0.21 & 10.1877 & +3.56 \\
        \bottomrule
    \end{tabularx}
    \caption{Actual training budgets at the largest endpoint of each tier. Token counts are in billions and $C$ is in units of $10^{20}$ FLOPs. Signed deviations from the same tier's MoE Base are computed before rounding: $\Delta D$ is in percent and $\Delta C$ in $10^{-4}\%$.}
    \label{tab:actual_training_budgets}
    \end{minipage}
\end{table}

We recompute the budgets from each run's frozen configuration and execution schedule. Independent formulas agree with the frozen budget utilities, and parameter counts obtained from the native model architectures agree with the saved checkpoint tensor inventories. We also check that actual training configurations, optimizer steps, and consumed tokens agree with the completed run records for all 45 scaling endpoints.

Table~\ref{tab:resource_deviations} reports the resource values and their deviations from MoE Base. For any quantity $X$, we define the signed percentage difference as
\begin{equation}
    \Delta X=100\left(\frac{X}{X_{\mathrm{MoE\ Base}}}-1\right),
    \label{eq:budget_deviation}
\end{equation}
using the baseline at the same tier and reference token budget. Differences are calculated before rounding the displayed values. Across all 45 endpoints, the maximum absolute deviations are 0.566\% for backbone parameters, 0.704\% for total parameters, 0.781\% for KV storage, and 0.901\% for FLOPs per step. Integer training step adjustments reduce the maximum total training FLOPs deviation to 0.00943\%. Table~\ref{tab:actual_training_budgets} gives the actual training budgets at the largest endpoint of each tier. Total computation includes the shared training prefix and the annealing branch, counting each prefix once for that endpoint.

For FLOPs, a multiply-add counts as two operations. We count attention projections, valid causal query--key and attention--value pairs, activated expert matrices, routers, the vocabulary projection, and the learned decoder operations. Sequence compression changes both projection costs and the number of attention pairs. Training FLOPs are estimated as three times the forward count. Other elementwise operations, optimizer updates, communication, padding overhead, and recomputation are excluded. KV storage counts BF16 keys and values across actual layer executions, including separate caches for repeated visits. It excludes the small fixed-size buffers used for incremental aggregation and writeback.

\section{Additional Implementation Details}
\label{sec:additional_implementation}

\paragraph{Architecture and data.}
We use a decoder-only Transformer with sparse MoE layers in the style of Qwen3~\citep{yang2025qwen3}. We fix the active expert count at eight per token, with normalized routing weights. The backbone uses RMSNorm~\citep{zhang2019root}, rotary position embeddings~\citep{su2024roformer}, grouped query attention~\citep{ainslie2023gqa} with query and key normalization, and SwiGLU experts~\citep{shazeer2020glu}. Attention uses eight query heads and two KV groups. The 1.3B, 3B, and 7B parameter tiers have 12, 16, and 20 backbone layers, respectively. All models use the same fixed training stream from FineWeb-Edu~\citep{penedo2024fineweb} sample350BT, with exact tokenized document deduplication and a separate 0.1B token validation set, tokenized with the 32K TinyLlama tokenizer~\citep{zhang2024tinyllama}. Training follows a warmup-stable-decay (WSD) learning rate schedule~\citep{hu2024minicpm}.

\paragraph{Training and validation.}
Table~\ref{tab:training_hyperparameters} lists the common training hyperparameters. The schedule describes the complete training trajectory: warmup is applied once when training starts from scratch, and annealing occupies the final 20\% of each endpoint's trajectory, rounded to complete optimizer updates. Gradient accumulation maintains the same global batch size across GPU allocations. Annealing branches inherit the model, optimizer, data cursor, and random states from their source checkpoint without restarting warmup. Every method consumes the same ordered training stream without wrapping; shorter budgets use shorter prefixes of this stream. Training and validation documents are separated before tokenization output is assembled into prediction windows. Each final validation evaluates all 24,415 fixed windows of length 4096, totaling 100,003,840 prediction tokens. We report token weighted cross-entropy without the router auxiliary loss. Any padding used to balance distributed validation batches is excluded from the loss numerator and token count.

\begin{table}[!t]
    \centering
    \small
    \setlength{\tabcolsep}{6pt}
    \renewcommand{\arraystretch}{1.12}
    \begin{tabularx}{\linewidth}{@{}Xr@{}}
        \toprule
        \multicolumn{1}{@{}c}{\bfseries\boldmath Hyperparameter} & \multicolumn{1}{c@{}}{\bfseries\boldmath Value} \\
        \midrule
        Training precision & BF16 mixed precision \\
        Context length & 4096 \\
        Global batch size & 256 sequences \\
        Prediction tokens per update & 1,048,576 \\
        Optimizer & AdamW~\citep{loshchilov2017decoupled} \\
        $(\beta_1,\beta_2)$ & $(0.9,0.95)$ \\
        Weight decay & 0.1 \\
        Gradient clipping norm & 1.0 \\
        Learning rate schedule & WSD \\
        Warmup & 500 optimizer updates \\
        Peak learning rate $\eta_{\max}$ & $10^{-3}$ \\
        Minimum learning rate $\eta_{\min}$ & $10^{-4}$ \\
        Final annealing fraction & 20\% \\
        Annealing rule & $\eta_{\min}+(\eta_{\max}-\eta_{\min})(1-\sqrt{p})$~\citep{hagele2024scaling} \\
        Training seed & 3407 \\
        Router auxiliary loss coefficient & $10^{-3}$ \\
        Linear/embedding initialization std & 0.02 \\
        RMSNorm $\epsilon$ & $10^{-6}$ \\
        \bottomrule
    \end{tabularx}
    \caption{Common training hyperparameters. Warmup is applied once per trajectory started from scratch. The variable $p\in[0,1]$ denotes progress through the annealing phase.}
    \label{tab:training_hyperparameters}
\end{table}

\paragraph{Scaling budgets and branches.}
\begin{table}[!htbp]
    \centering
    \begin{minipage}{0.58\textwidth}
    \centering
    \small
    \setlength{\tabcolsep}{4pt}
    \begin{tabular}{@{}llr@{}}
        \toprule
        {\bfseries\boldmath Size} & {\bfseries\boldmath Token budgets (B)} & {\bfseries\boldmath Points} \\
        \midrule
        1.3B & 5, 10, 20, 40 & 4 \\
        3B & 5, 10, 20, 40, 70 & 5 \\
        7B & 5, 10, 20, 40, 70, 140 & 6 \\
        \bottomrule
    \end{tabular}
    \caption{Reference token budgets.}
    \label{tab:scaling_budgets}
\end{minipage}
\end{table}
Table~\ref{tab:scaling_budgets} lists the four, five, and six reference token budgets for the 1.3B, 3B, and 7B tiers, giving 15 validation NLL endpoints per architecture. These budgets are labeled by the MoE Base token count; the actual token counts of other architectures are adjusted to match total FLOPs.
For each architecture and model tier, we share one mainline across token budgets. We precompute the stable phase endpoint at 80\% of each target token budget, rounded down to a complete optimizer update. At each such point, the mainline saves a full checkpoint containing the model, optimizer, data cursor, and random states. A branch resumes from this checkpoint and anneals over the remaining budget, while the mainline continues at constant learning rate for larger endpoints. Branches inherit training progress and do not repeat warmup. The computation reported for each endpoint includes its shared training prefix and its own annealing branch.

\paragraph{Convolution encoder.}
In the convolution encoder variant, we replace the fixed mean with a causal convolution applied independently to each channel, using stride $k$:
\begin{equation}
    \begin{aligned}
        \operatorname{Enc}(H)_j
        &=\sum_{r\in\mathcal R} a_r\odot H_{kj-r},\\
        &\qquad 1\leq j\leq l.
    \end{aligned}
    \label{eq:learned_encoder}
\end{equation}
Here $\mathcal R$ is the set of nonnegative offsets in the causal kernel, $a_r\in\mathbb{R}^d$ contains the learnable coefficients for each channel, and $\odot$ denotes elementwise multiplication. States before the sequence start are set to zero. We initialize the coefficients to reproduce mean aggregation. Learning these coefficients allows the model to adjust both the local aggregation weights and the scale of the latent states.

\paragraph{Decoder parameterization.}
The effective coefficients of the scaled decoder are implemented as $w_s=1+\delta_s$, where each $\delta_s\in\mathbb{R}^d$ starts at zero. Thus, initialization exactly recovers direct writeback. AdamW optimizes the correction parameters $\delta_s$, including weight decay on these parameters. The coefficients are unconstrained and shared across token groups. We report the effective coefficients $w_s$.

\paragraph{Causal incremental decoding.}
The latent cache is indexed by completed token groups, while RoPE uses the zero-based position ID $2j-1$ of each group's final token, whose one-based index is $2j$. During cached decoding, an unfinished pair is retained until its second token arrives. The completed pair is encoded and processed by the first visit, producing the increment written to positions $2j$ and $2j+1$. The second receiver uses the stored decoded increment when the next token arrives. The second visit uses the full sequence cache. Caches are indexed by layer execution, so visits never overwrite each other's keys or values even when weights are shared. Receivers outside the sequence are omitted, and incomplete groups generate no latent update until completed. All caches and aggregation/writeback buffers are reset for a new sequence.

\paragraph{Downstream evaluation.}
\begin{table}[!htbp]
    \centering
    \begin{minipage}{0.72\textwidth}
    \centering
    \small
    \setlength{\tabcolsep}{4pt}
    \renewcommand{\arraystretch}{1.12}
    \begin{tabularx}{\linewidth}{@{}>{\raggedright\arraybackslash}Xlc@{}}
        \toprule
        \multicolumn{1}{@{}c}{\bfseries\boldmath Benchmark} & \multicolumn{1}{c}{\bfseries\boldmath Harness task} & \multicolumn{1}{c@{}}{\bfseries\boldmath Metric} \\
        \midrule
        LAMBADA OpenAI & \texttt{lambada\_openai} & \texttt{acc} \\
        LAMBADA Standard & \texttt{lambada\_standard} & \texttt{acc} \\
        HellaSwag & \texttt{hellaswag} & \texttt{acc\_norm} \\
        PIQA & \texttt{piqa} & \texttt{acc\_norm} \\
        WinoGrande & \texttt{winogrande} & \texttt{acc} \\
        ARC-Easy & \texttt{arc\_easy} & \texttt{acc\_norm} \\
        ARC-Challenge & \texttt{arc\_challenge} & \texttt{acc\_norm} \\
        SciQ & \texttt{sciq} & \texttt{acc\_norm} \\
        MMLU continuation & \texttt{mmlu\_continuation} & \texttt{acc} \\
        \bottomrule
    \end{tabularx}
    \caption{Downstream task identifiers and scoring metrics. Both shot settings use the same metric. \texttt{acc\_norm} is the harness length-normalized accuracy.}
    \label{tab:downstream_protocol}
\end{minipage}
\end{table}
We use \texttt{lm-evaluation-harness}~\citep{eval-harness} version 0.4.13, with upstream commit \texttt{ddd6722}, through an adapter that loads each checkpoint's native model implementation. Table~\ref{tab:downstream_protocol} lists the task identifiers and reported metrics. We evaluate every task at both 0-shot and 5-shot, using few-shot seed 1234 and one evaluation per setting. No chat template or special prefix token is added to normal prompts. Task data, templates, few-shot sampling, and scoring configurations are identical across methods. For LAMBADA OpenAI at 5-shot, we explicitly select the original test example pool and exclude the current question. Other tasks retain their default example pools. When an input exceeds the context limit, we truncate the context from the left while retaining the complete scored continuation. Model weights are loaded in FP32, forward computation uses BF16 autocast, cross-entropy uses FP32 logits, and continuation log probabilities are summed in FP64. The reported average gives equal weight to the nine benchmark scores; MMLU continuation uses the harness aggregate accuracy.

\begin{figure}[!htbp]
    \centering
    \begin{minipage}{0.48\textwidth}
    \centering
    \includegraphics[width=\linewidth]{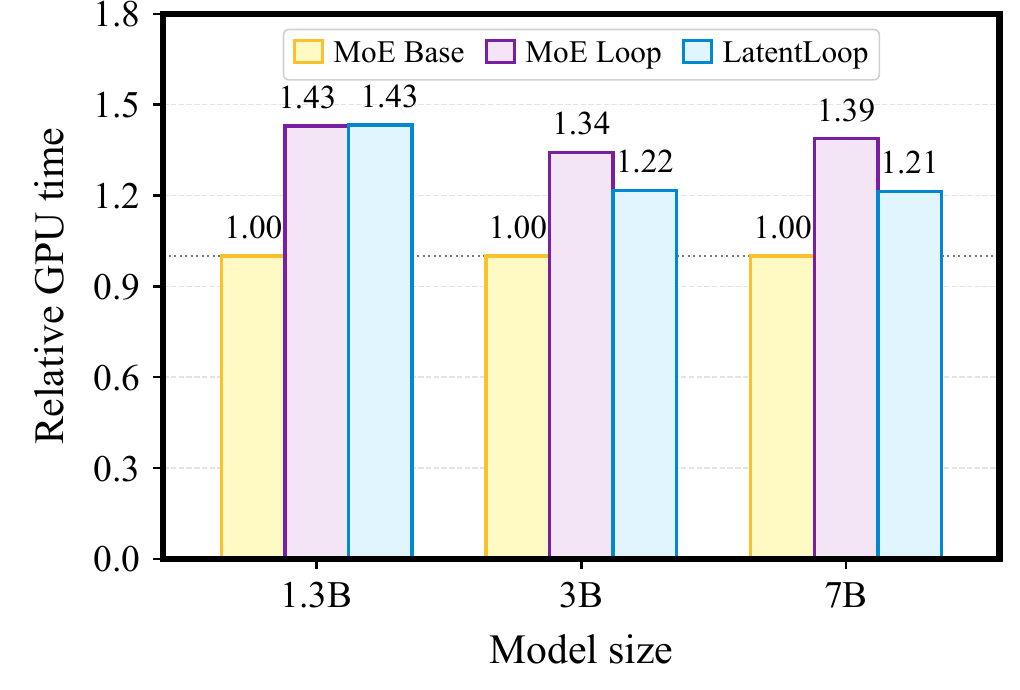}
    \caption{GPU time normalized to MoE Base at each model size. GPU time is elapsed training time multiplied by GPU count.}
    \label{fig:training_time}
    \end{minipage}
\end{figure}

\paragraph{Training time.}
Despite matched theoretical FLOPs, LatentLoop uses approximately 9\% and 13\% less GPU time than MoE Loop at 3B and 7B, respectively, with similar GPU time at 1.3B (Figure~\ref{fig:training_time}). By reallocating the saved first visit computation to model width, LatentLoop is wider than MoE Loop throughout the backbone. The larger matrix operations can improve hardware utilization, providing a possible explanation for the lower training cost.
Actual training speed depends on how efficiently the implementation uses the available hardware.

\paragraph{Training time measurement.}
Figure~\ref{fig:training_time} uses the longest stable mainline for each method and tier. We multiply elapsed time by GPU count and normalize by MoE Base at the same tier. The timer starts immediately before the optimizer loop and includes validation and checkpoint writes through completion. It excludes initialization, qualification, queue waits, and post-run checkpoint auditing. Resumed mainlines combine the original prefix up to the resume point with the continuation, excluding discarded steps; a resume boundary between log samples uses interpolation of cumulative elapsed time. The 1.3B baselines use eight GPUs while LatentLoop uses 64; all 3B and 7B runs use 64 GPUs. The figure therefore measures GPU resource use, with execution dates and pools also varying across runs. Completing the full 7B scaling ladder for LatentLoop required about 35 hours on 64 NVIDIA H200 GPUs. This duration sums the active runs of its shared mainline and six annealing branches, including their qualification runs.

\section{Additional Ablations}
\label{sec:additional_ablations}

These ablations use the 3B/40B reference setting and follow the training and budget matching protocols of the main ablations in Section~\ref{sec:ablation_studies}.

\par

\begin{table}[!htbp]
    \centering
    \begin{minipage}[t]{0.48\textwidth}
    \vspace{0pt}
    \centering
\small
\renewcommand{\arraystretch}{1.12}
\setlength{\tabcolsep}{4pt}
\begin{tabularx}{\linewidth}{@{}Xcc@{}}
    \toprule
     & \multicolumn{2}{c@{}}{\bfseries\boldmath MoE} \\
    \cmidrule(lr){2-3}
    {\bfseries\boldmath Attention} & {\bfseries\boldmath Shared} & {\bfseries\boldmath Untied} \\
    \midrule
    Shared & 2.0237 & 2.0348 \\
    Untied & \cellcolor{gray!15}\textbf{2.0199} & 2.0340 \\
    \midrule
    \multicolumn{3}{@{}l@{}}{\textit{References}} \\
    \multicolumn{2}{@{}l}{MoE Base} & 2.0426 \\
    \multicolumn{2}{@{}l}{MoE Loop} & 2.0293 \\
    \bottomrule
\end{tabularx}
\caption{Attention and MoE sharing (validation NLL).}
\label{tab:attn_moe_sharing}

\end{minipage}\hfill
    \begin{minipage}[t]{0.48\textwidth}
    \vspace{0pt}
    \centering
\small
\renewcommand{\arraystretch}{1.12}
\setlength{\tabcolsep}{4pt}
\begin{tabularx}{\linewidth}{@{}Xcc@{}}
    \toprule
     & \multicolumn{2}{c@{}}{\bfseries\boldmath Second visit} \\
    \cmidrule(lr){2-3}
    {\bfseries\boldmath First visit} & {\bfseries\boldmath Full} & {\bfseries\boldmath Latent} \\
    \midrule
    Full & 2.0293 & 2.0386 \\
    Latent & \cellcolor{gray!15}\textbf{2.0199} & 2.0467 \\
    \bottomrule
\end{tabularx}
\caption{Validation NLL by latent position. Full/full is MoE Loop; the three latent variants use independent attention blocks.}
\label{tab:latent_position}

\end{minipage}
\end{table}

\paragraph{Attention and MoE sharing.}
Table~\ref{tab:attn_moe_sharing} compares the four attention and MoE sharing choices, using the final LatentLoop configuration for shared MoE and independent attention. All use the same mean encoder and learned writeback, with model width and attention dimensions fixed; expert counts and training steps are adjusted to match the budgets. LatentLoop gives the lowest NLL, at 2.0199, compared with 2.0237 for full sharing. Untying MoE gives higher NLL whether attention is shared (2.0348) or independent (2.0340). Separate MoE routers and expert weights consume much more of the parameter budget: matching it reduces the expert count per pool from 76 to 51 with shared attention, and from 75 to 50 with independent attention. These results favor allocating the parameter budget to larger expert pools reused across visits, while giving attention independent projections at a small additional parameter cost.
\par

\par

\paragraph{Latent position.}
Table~\ref{tab:latent_position} compares the four choices of full or latent resolution across the two visits, using MoE Loop as the full/full reference. The three configurations with compression share MoE routers and experts while retaining independent attention blocks. Each compressed visit uses $k=2$, mean aggregation, and the scaled decoder. When both visits are compressed, each independently encodes the current full sequence and writes its increment back before the next visit; the two writebacks share the same writeback coefficients.

Compressing only the first visit gives the lowest NLL, at 2.0199. Moving compression to the second visit increases NLL to 2.0386, despite identical parameter counts, KV storage, FLOPs per step, and training updates. Compressing both visits gives 2.0467, even though budget matching increases model width from 1296 to 1440. Thus, the additional width does not compensate for compressing the second visit in this setting. The full/full control with independent attention blocks also outperforms both variants that compress the second visit, at 2.0319 (Table~\ref{tab:ablation_sharing}). These comparisons support our asymmetric allocation: reducing computation in the first visit while retaining full resolution in the second.
\par

\par

\begin{table}[!htbp]
    \centering
    \begin{minipage}[t]{0.48\textwidth}
    \vspace{0pt}
    \centering
    \small
\renewcommand{\arraystretch}{1.12}
\setlength{\tabcolsep}{3pt}
\begin{tabularx}{\linewidth}{@{}Xr@{}}
    \toprule
    {\bfseries\boldmath Configuration} & {\bfseries\boldmath NLL $\downarrow$} \\
    \midrule
    \multicolumn{2}{@{}l}{\textit{Baselines}} \\
    MoE Base & 2.0426 \\
    MoE Loop & 2.0293 \\
    \midrule
    \multicolumn{2}{@{}l}{\textit{LatentLoop variants}} \\
    \rowcolor{gray!15}
    $k=2$ (ours) & \textbf{2.0199} \\
    $k=3$ & 2.0257 \\
    $k=4$ & 2.0297 \\
    $k=2$, one token delayed writeback & 2.0269 \\
    \bottomrule
\end{tabularx}
\caption{Compression factor and writeback delay.}
\label{tab:compression_delay}

\end{minipage}\hfill
    \begin{minipage}[t]{0.48\textwidth}
    \vspace{0pt}
    \captionsetup{type=figure}
    \centering
    \includegraphics[width=\linewidth]{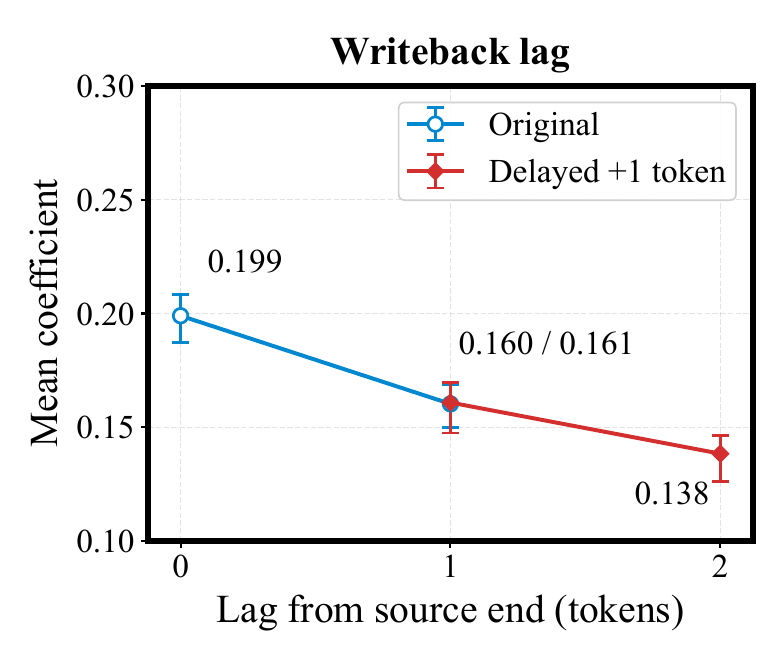}
    \caption{Final writeback coefficient means in the original and delayed models, aligned by distance from the source group's last token. Whiskers show channel 25th--75th percentiles.}
    \label{fig:delayed_writeback_coefficients}
\end{minipage}
\end{table}

\paragraph{Compression factor and writeback delay.}
We next test $k=3$ and $k=4$ with the same two loop visits and independently matched budgets. When $k$ does not divide the sequence length, only complete groups produce latent states; the remaining tokens are retained in the full sequence and can receive updates from the preceding group. Table~\ref{tab:compression_delay} shows higher NLLs of 2.0257 and 2.0297, compared with 2.0199 for $k=2$. We believe increased writeback delay is a major reason: a latent update can only be used after its source group is complete, and larger groups increase the gap between later receiver positions and the most recent token included in their update. To test this explanation, we retain $k=2$ and delay both receiver positions by one token, from $(2j,2j+1)$ to $(2j+1,2j+2)$, with no change to the encoder, decoder design, or compute budget. NLL rises to 2.0269, supporting the importance of timely writeback even without stronger compression. This observation is consistent with SpiralFormer, which reports worse performance when an additional one token shift removes the overlap between source and receiver groups~\citep{yu2026spiralformer}. Its Appendix E also discusses the resulting lag behind the full resolution stream. 
\par

Figure~\ref{fig:delayed_writeback_coefficients} compares the final coefficients of the original and delayed models. Delaying both receivers by one token lowers their channel means to 0.161 and 0.138, approximately 19\% and 14\% below the original values. In particular, the delayed model's $w_0$ mean of 0.161 is close to the original model's $w_1$ mean of 0.160. Both receivers lie one token after the source pair's final token, despite occupying different decoder slots. This agreement links the learned writeback strength to how recent the information in the latent update is: updates from more distant context receive smaller weights. It provides further support for our interpretation that delayed updates are less well aligned with the receiver's current context.

\section{Comparison with SpiralFormer}
\label{sec:spiralformer_comparison}

\begin{table}[!htbp]
    \centering
    \small
    \setlength{\tabcolsep}{3pt}
    \renewcommand{\arraystretch}{1.12}
    \resizebox{\textwidth}{!}{%
    \begin{tabular}{@{}cccl*{10}{c}@{}}
        \toprule
        \multicolumn{1}{@{}c}{\bfseries\boldmath Shot} & \multicolumn{1}{c}{\bfseries\boldmath Params} & \multicolumn{1}{c}{\bfseries\boldmath Token} & \multicolumn{1}{c}{\bfseries\boldmath Model} & \multicolumn{1}{c}{\bfseries\boldmath LD-O} & \multicolumn{1}{c}{\bfseries\boldmath LD-S} & \multicolumn{1}{c}{\bfseries\boldmath HS} & \multicolumn{1}{c}{\bfseries\boldmath PIQA} & \multicolumn{1}{c}{\bfseries\boldmath WG} & \multicolumn{1}{c}{\bfseries\boldmath ARC-E} & \multicolumn{1}{c}{\bfseries\boldmath ARC-C} & \multicolumn{1}{c}{\bfseries\boldmath SciQ} & \multicolumn{1}{c}{\bfseries\boldmath MMLU} & \multicolumn{1}{c@{}}{\bfseries\boldmath Avg} \\
        \midrule
        \raisebox{\dimexpr-\arraystretch\ht\strutbox/2-\arraystretch\dp\strutbox/2\relax}[0pt][0pt]{5-shot} & 1.4B / 1.4B & 250B & SpiralFormer-L & 46.32 & 45.29 & 51.66 & 70.51 & 59.51 & 60.61 & 31.91 & 93.20 & 30.35 & 54.37 \\
         & 7B / 1B & 140B & \cellcolor{gray!15}LatentLoop & \cellcolor{gray!15}\textbf{49.21} & \cellcolor{gray!15}\textbf{45.92} & \cellcolor{gray!15}\textbf{68.44} & \cellcolor{gray!15}\textbf{76.33} & \cellcolor{gray!15}\textbf{63.93} & \cellcolor{gray!15}\textbf{79.55} & \cellcolor{gray!15}\textbf{51.71} & \cellcolor{gray!15}\textbf{96.20} & \cellcolor{gray!15}\textbf{39.35} & \cellcolor{gray!15}\textbf{63.40} \\
        \midrule
        \raisebox{\dimexpr-\arraystretch\ht\strutbox/2-\arraystretch\dp\strutbox/2\relax}[0pt][0pt]{0-shot} & 1.4B / 1.4B & 250B & SpiralFormer-L & 52.86 & \textbf{46.69} & 50.88 & 69.64 & 56.43 & 50.59 & 28.75 & 80.70 & 29.24 & 51.75 \\
         & 7B / 1B & 140B & \cellcolor{gray!15}LatentLoop & \cellcolor{gray!15}\textbf{54.92} & \cellcolor{gray!15}45.22 & \cellcolor{gray!15}\textbf{66.91} & \cellcolor{gray!15}\textbf{75.68} & \cellcolor{gray!15}\textbf{62.59} & \cellcolor{gray!15}\textbf{74.20} & \cellcolor{gray!15}\textbf{46.50} & \cellcolor{gray!15}\textbf{90.40} & \cellcolor{gray!15}\textbf{37.69} & \cellcolor{gray!15}\textbf{61.57} \\
        \bottomrule
    \end{tabular}%
    }
    \caption{System comparison on downstream benchmarks (\%). Params denote approximate total/active parameter counts. Benchmark abbreviations follow Table~\ref{tab:downstream}; Avg is the unweighted mean of the nine scores.}
    \label{tab:spiralformer_system}
\end{table}

We compare our largest LatentLoop model against the largest dense SpiralFormer-L model. Both studies use \texttt{lm-evaluation-harness}~\citep{eval-harness} to evaluate the same nine benchmarks under 0-shot and 5-shot settings. Because SpiralFormer's implementation was not publicly available, we quote the task scores reported in Appendix H, Table 3 of \citet{yu2026spiralformer}. Table~\ref{tab:spiralformer_system} presents the system comparison. The systems use different training data, tokenizers, training budgets, and other implementation details, which can also contribute to the observed differences.

\section{Full Scaling Results}
\label{sec:full_scaling_results}

Table~\ref{tab:scaling_endpoints} reports validation NLL across the 45 model and budget configurations considered in Section~\ref{sec:scaling_analysis}. Each value is measured on the same fixed 0.1B token validation set. Token budgets are reference budgets; actual token counts are adjusted to match total training FLOPs across methods. LatentLoop achieves the lowest NLL at all 15 matched budgets.

\begin{table}[!htbp]
    \centering
    \begin{minipage}{0.68\linewidth}
    \centering
    \small
    \setlength{\tabcolsep}{3pt}
    \renewcommand{\arraystretch}{1.1}
    \begin{tabular}{@{}ccccc@{}}
        \toprule
        {\bfseries\boldmath Params} & {\bfseries\boldmath Tokens} & {\bfseries\boldmath MoE Base} & {\bfseries\boldmath MoE Loop} & {\bfseries\boldmath LatentLoop} \\
        \midrule
        1.3B & 5B & 2.4093 & 2.4105 & \textbf{2.3979} \\
         & 10B & 2.3026 & 2.3004 & \textbf{2.2863} \\
         & 20B & 2.2201 & 2.2161 & \textbf{2.2022} \\
         & 40B & 2.1572 & 2.1519 & \textbf{2.1393} \\
        \midrule
        3B & 5B & 2.3157 & 2.3077 & \textbf{2.2971} \\
         & 10B & 2.2010 & 2.1910 & \textbf{2.1805} \\
         & 20B & 2.1107 & 2.0988 & \textbf{2.0894} \\
         & 40B & 2.0426 & 2.0293 & \textbf{2.0199} \\
         & 70B & 1.9977 & 1.9837 & \textbf{1.9771} \\
        \midrule
        7B & 5B & 2.2561 & 2.2425 & \textbf{2.2336} \\
         & 10B & 2.1351 & 2.1217 & \textbf{2.1117} \\
         & 20B & 2.0377 & 2.0250 & \textbf{2.0156} \\
         & 40B & 1.9634 & 1.9510 & \textbf{1.9429} \\
         & 70B & 1.9135 & 1.9016 & \textbf{1.8950} \\
         & 140B & 1.8640 & 1.8546 & \textbf{1.8482} \\
        \bottomrule
    \end{tabular}
    \caption{Validation NLL ($\downarrow$) at all scaling endpoints. Parameter tiers denote approximate total parameter counts. The lowest NLL at each matched budget is bold.}
    \label{tab:scaling_endpoints}
    \end{minipage}
\end{table}

\end{document}